%% file: main.tex
\documentclass{article}
\usepackage{iclr2025_conference,times}
\input{math_commands.tex} 
\usepackage{hyperref}
\usepackage{url}
\usepackage{graphicx}
\usepackage{booktabs}
\usepackage{amsmath,amssymb}
\usepackage{microtype}
\usepackage{enumitem}
\usepackage{tikz}
\usepackage{comment}
\usepackage{booktabs}
\usepackage{subcaption}
\usepackage{placeins}

\usetikzlibrary{arrows.meta,positioning,fit,backgrounds,calc,shapes.misc}

\usepackage{booktabs,tabularx}
\usepackage[utf8]{inputenc} 
\usepackage{pifont}
\usepackage{placeins}
\newcommand{\statusFull}{\ding{51}}     
\newcommand{\statusPartial}{\ding{72}}  
\newcommand{\statusNone}{\ding{109}}    
\setcitestyle{yysep={,}}

\iclrfinalcopy 

\hypersetup{
  pdftitle={Generalist Open-World Temporal Perception},
  pdfauthor={Cristian Sminchisescu},
  pdfsubject={Position paper on generalist open-world temporal perception},
  pdfkeywords={temporal foundation models, multimodal perception,
    open-world perception, world models, physical AI}
}

\hypersetup{hidelinks}
\title{Generalist Open-World Temporal Perception}
\author{Cristian Sminchisescu$^{1,2,3}$\thanks{This position paper builds on published work carried out primarily while at Google DeepMind. The perspectives, arguments, interpretations, and conclusions presented here are entirely those of the author and do not represent an institutional position.}\\
\text{$^1$Google DeepMind, $^2$Lund University, $^3$IMAR}
}

\begin{document}

\maketitle
\thispagestyle{plain}

\begin{abstract}

The next generation of artificial intelligence systems will likely be natively temporal and multimodal in both inputs and outputs: able to converse, perceive, predict, reason, and synthesize through a shared representation of the world. 
Realizing this requires more than attaching image, audio, or video to a primarily textual model. 
It requires a temporal perceptual substrate in which sensory streams, language, and structured outputs are integrated within a common multimodal world model. We seek a generalist open-world perceptual system that represents biological forms, natural physical structures, and artifacts, and their interactions, as a coherent, temporally persistent process. 
The model should infer geometry, articulation, semantics, interaction structure, and uncertainty from raw multimodal streams; maintain identity through occlusion and viewpoint change; generalize across species, forms, mechanisms, and materials; and abstain or expand its ontology when encountering the unknown. 
The objective is to construct a structured world state that supports understanding, prediction, counterfactual reasoning, and controllable synthesis. Recent work on large video models suggests that some cross-modal and reasoning-like capabilities can emerge from large-scale generative video pretraining in a manner reminiscent of language-model scaling.  Yet these capabilities are often accessed through language probes or expressed through photorealistic video, leaving explicit semantic, geometric, or temporal structure largely unexposed. 
This paper articulates an alternative and complementary paradigm: perception and synthesis can be framed as distinct conditionings within a shared Generalist Open-World Temporal Perception Architecture (GOWTPA).
By incorporating synthesis principles into perception, such models may provide a foundation for open-world, multimodal, and temporally coherent understanding, where recognition, structured prediction, and simulation arise as different conditionings of the same generative substrate, while reasoning and embodiment-specific policies build upon the resulting world state. This positions generalist temporal perception as a potential foundation layer for broader multimodal intelligence and physical AI.

\end{abstract}

\section{Introduction}

Understanding the world as a coherent, dynamic, and structured process remains a central challenge for artificial intelligence. Despite major progress in recognition, generation, and reasoning, today’s perceptual systems do not yet reliably maintain a model of reality that generalizes beyond training distributions, persists through time, incorporates newly discovered structure, and adapts to novel situations without exhaustive supervision.

While large language and vision--language models have shown that scaling and self-supervision can unlock broad reasoning and transfer capacities~\citep{brown2020language,alayrac2022flamingo}, comparable generality has not yet been established for open-world perception. This requires inferring geometry, semantics, articulation, interactions, and predictive structure from multimodal sensory streams across the diversity and complexity of the open world.

The central gap is no longer the absence of strong perceptual components. 
The field now has powerful visual encoders, open-vocabulary models, promptable segmentation systems, dense geometric predictors, video--language models, and increasingly capable video generators. 
What remains missing is a common substrate that turns these components into explicit world-state inference. 
Existing systems often expose isolated projections of sensory input; perception requires those projections to be coupled through persistent entities, multimodal evidence, uncertainty, and temporal memory.

A growing body of work suggests that the generative route may provide a
more unified foundation.  When diffusion, flow-matching, or autoregressive transformers are trained to model temporally coherent video, they appear to internalize motion, persistence, articulation, and physical regularities as part of modeling the video distribution. Recent work has begun to examine whether large video generative models acquire reusable visual priors and reasoning-like abilities from large-scale generative video pretraining~\citep{acuaviva2025generation,wiedemer2025video}, a question closely related to our work on generalist open-world perceptual intelligence. 
Here the objective is to expose the temporal structure learned by generative video models as explicit semantic, geometric, and relational perceptual state. A model may internalize useful regularities of motion, persistence, and scene evolution while leaving the physical factors that explain the scene entangled or inaccessible to control.

A useful starting point is the classical conditional generative view of perception, consistent with analysis-by-synthesis, Bayesian, and predictive-coding accounts of vision~\citep{kersten2003bayesian,yuille2006vision,rao1999predictive}. 
For a long time, perception has been understood as inference under a model, with parameters  $\mathbf{\theta}$,  of latent world states $\mathbf{s}$ and sensory observations $\mathbf{o}$: synthesis corresponds to sampling or predicting observations given latent causes, while perception corresponds to inferring world states from observations. In probabilistic terms, synthesis concerns distributions such as $p_\mathbf{\theta}(\mathbf{o}\mid\mathbf{s})$, whereas perception concerns posterior inference over world state, $p_\mathbf{\theta}(\mathbf{s}\mid\mathbf{o})$, possibly conditioned, in addition, on task, modality, or context. The present setting transposes this classical generative view to temporal foundation models trained on large video and multimodal corpora. We argue that under a suitable tokenized representation, synthesis and structured perception become complementary conditionings of a shared temporal world model, differing in initialization, conditioning, inference or sampling regime, parameter specialization, and readout. The central question is whether these backbones can expose persistent, explicit, and open-world perceptual structure.

Motivated by several empirical trends, we formulate four hypotheses.
First, generative video and audio--visual backbones appear to contain latent spatio-temporal structure that can be exposed through structured readouts. 
Second, temporal coherence and cross-modal correlation provide cues for entity formation without requiring all entities to be labeled in advance. 
Third, articulation and correspondence priors may transfer across related biological, mechanical, and artificial systems. 
Fourth, jointly predicting correlated structured outputs can align the latent representation and expose signals of predictive reliability.

In this paper, by \emph{world state} we do not mean a complete physical simulator state or a single privileged representation of reality. Here, we envision a temporally persistent, queryable representation that binds sensory evidence to explicit perceptual variables, including geometry, semantics, dynamics, relations, and modality-specific projections such as language descriptions or sound-source estimates. The state may be partial, approximate, task-dependent, and probabilistic. Its existence is tested through persistence across missing observations, compatibility among structured readouts, revision under new evidence, and predictive validity under changes in time, viewpoint, or modality. The \emph{world model} denotes the learned representations and dynamics through which such states are inferred, predicted, and decoded. 

\usetikzlibrary{positioning,calc}

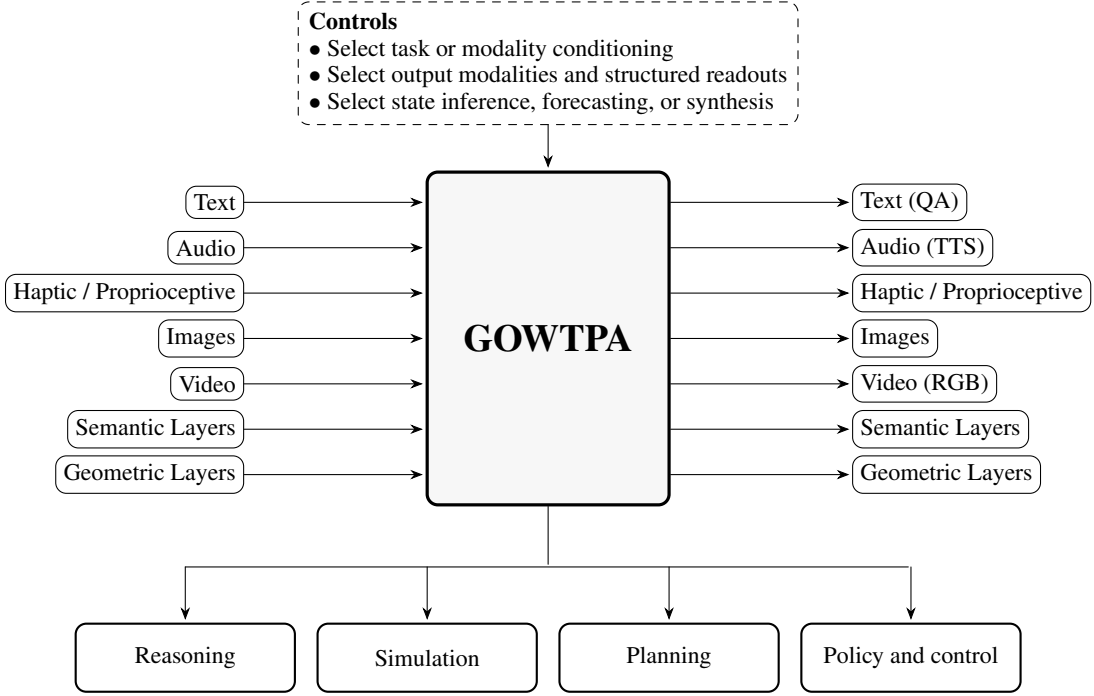
\begin{figure}[t]
\centering
\begin{tikzpicture}[
  font=\small,
  box/.style={
    draw,
    fill=black!3,
    rounded corners,
    inner sep=5pt,
    align=center,
    minimum width=32mm,
    minimum height=44mm,
    very thick
  },
  io/.style={
    draw,
    rounded corners,
    inner sep=3pt,    align=center
  },
  ctrl/.style={
    draw,
    dashed,
    rounded corners,
    inner sep=4pt,
    align=left
  },
  downstream/.style={
    draw,
    rounded corners,
    inner sep=4pt,
    align=center,
    font=\footnotesize,
    minimum width=29mm,
    minimum height=9mm,
    thick
  },
  >=Stealth,
  line cap=round,
  shorten >=1pt
]

\node[box] (core) {\Large\textbf{GOWTPA}};

\foreach \i/\dy in {
  1/18mm,
  2/12mm,
  3/6mm,
  4/0mm,
  5/-6mm,
  6/-12mm,
  7/-18mm
}{
  \coordinate (L\i) at ($(core.west)+(0,\dy)$);
  \coordinate (R\i) at ($(core.east)+(0,\dy)$);
}

\node[io, anchor=east] (txt) at ($(L1)+(-24mm,0)$)
  {Text};
\node[io, anchor=east] (aud) at ($(L2)+(-24mm,0)$)
  {Audio};
\node[io, anchor=east] (hap) at ($(L3)+(-24mm,0)$)
  {Haptic / Proprioceptive};
\node[io, anchor=east] (img) at ($(L4)+(-24mm,0)$)
  {Images};
\node[io, anchor=east] (vid) at ($(L5)+(-24mm,0)$)
  {Video};
\node[io, anchor=east] (sem) at ($(L6)+(-24mm,0)$)
  {Semantic Layers};
\node[io, anchor=east] (geo) at ($(L7)+(-24mm,0)$)
  {Geometric Layers};

\node[io, anchor=west] (txto) at ($(R1)+(24mm,0)$)
  {Text (QA)};
\node[io, anchor=west] (audo) at ($(R2)+(24mm,0)$)
  {Audio (TTS)};
\node[io, anchor=west] (hapo) at ($(R3)+(24mm,0)$)
  {Haptic / Proprioceptive};
\node[io, anchor=west] (imgo) at ($(R4)+(24mm,0)$)
  {Images};
\node[io, anchor=west] (vido) at ($(R5)+(24mm,0)$)
  {Video (RGB)};
\node[io, anchor=west] (semo) at ($(R6)+(24mm,0)$)
  {Semantic Layers};
\node[io, anchor=west] (geoo) at ($(R7)+(24mm,0)$)
  {Geometric Layers};

\draw[->] (txt.east) -- ++(4.8mm,0) |- (core.west |- L1);
\draw[->] (aud.east) -- ++(4.8mm,0) |- (core.west |- L2);
\draw[->] (hap.east) -- ++(4.8mm,0) |- (core.west |- L3);
\draw[->] (img.east) -- ++(4.8mm,0) |- (core.west |- L4);
\draw[->] (vid.east) -- ++(4.8mm,0) |- (core.west |- L5);
\draw[->] (sem.east) -- ++(4.8mm,0) |- (core.west |- L6);
\draw[->] (geo.east) -- ++(4.8mm,0) |- (core.west |- L7);

\draw[->] (core.east |- R1) -- ++(4.8mm,0) |- (txto.west);
\draw[->] (core.east |- R2) -- ++(4.8mm,0) |- (audo.west);
\draw[->] (core.east |- R3) -- ++(4.8mm,0) |- (hapo.west);
\draw[->] (core.east |- R4) -- ++(4.8mm,0) |- (imgo.west);
\draw[->] (core.east |- R5) -- ++(4.8mm,0) |- (vido.west);
\draw[->] (core.east |- R6) -- ++(4.8mm,0) |- (semo.west);
\draw[->] (core.east |- R7) -- ++(4.8mm,0) |- (geoo.west);

\node[ctrl, above=6mm of core.north] (ctrls) {
  \textbf{Controls}\\
  \(\bullet\) Select task or modality conditioning\\
  \(\bullet\) Select output modalities and structured readouts\\
  \(\bullet\) Select state inference, forecasting, or synthesis
};
\draw[->] (ctrls.south) -- (core.north);

\node[downstream] (reasoning)
  at ($(core.south)+(-48mm,-20mm)$)
  {Reasoning};

\node[downstream] (simulation)
  at ($(core.south)+(-16mm,-20mm)$)
  {Simulation};

\node[downstream] (planning)
  at ($(core.south)+(16mm,-20mm)$)
  {Planning};

\node[downstream] (policy)
  at ($(core.south)+(48mm,-20mm)$)
  {Policy and control};

\coordinate (bus) at ($(core.south)+(0,-8mm)$);
\draw[-] (core.south) -- (bus);
\draw[-] (reasoning.north |- bus) -- (policy.north |- bus);
\draw[->] (reasoning.north |- bus) -- (reasoning.north);
\draw[->] (simulation.north |- bus) -- (simulation.north);
\draw[->] (planning.north |- bus) -- (planning.north);
\draw[->] (policy.north |- bus) -- (policy.north);

\end{tikzpicture}

\caption{
\textbf{Generalist Open-World Temporal Perception Architecture.}
GOWTPA is a conceptual multimodal architecture for investigating perception and
generation as complementary conditionings of shared temporal perceptual state.
Different instantiations may condition on, and predict, subsets of text,
audio, haptic or proprioceptive signals, images, video, and structured
semantic or geometric variables. Controls select the conditioning sources,
requested outputs, and operating mode. The resulting perceptual state and
readouts can support downstream reasoning, simulation, planning, and policy
or control.
}
\label{fig:gowtpa}
\end{figure}

\paragraph{Scope and aim.} This position paper characterizes the computational requirements of generalist open-world temporal perception (\emph{what}) and develops conditional generative temporal models as one plausible route toward satisfying them (\emph{how}). The computational requirements should remain comparatively stable, while architectures, representations, objectives, and training regimes will continue to evolve.

\paragraph{Approach and levels of description.}
The proposal is more specific than the general claim that intelligent perceptual systems require world models. Computationally, it calls for a persistent and revisable temporal state, queryable through mutually consistent predictions of geometry, semantics, entities, and events, together with their uncertainty. Algorithmically, conditional generative perception provides one route toward learning and exposing such a state. At the implementation level, contemporary temporal generative backbones offer promising, but non-exclusive, substrates for investigating this route, with THFM and GenCeption providing initial empirical support for components of the approach \citep{wang2026thfm,wang2026eccv}.

\paragraph{Strategic implication for physical AI.} Physical AI is commonly framed around systems that perceive, reason, and act in the physical world \citep{nvidia2025cosmos, abdolmaleki2025geminirobotics15}. Our proposal is to treat explicit temporal world state as a reusable foundation layer between multimodal sensing and task-specific control. The shared perceptual core represents entities, geometry, relations, events, and uncertainty, while sensors, readouts, affordances, and policies remain specialized to each embodiment and application. This shifts the top-down program from repeatedly constructing product-specific perception stacks towards developing a common perceptual platform that can support humanoid robots and other autonomous systems, assistive devices, scientific instruments, spatial computing, and controllable synthesis.

\section{Layers of Perceptual Agency}
\label{sec:levels-of-observation}

Perception and agency can be distinguished by whether observation is treated as
given or as a decision variable. This distinction is consistent with classical
work on active perception, active vision, and animate vision, in which sensing
is shaped by task and the system may control how evidence is acquired
\citep{bajcsy1988active,aloimonos1988active,ballard1991animate}.

At the perceptual level, a system receives multimodal sensory evidence and
infers a structured representation of the external world. Its task is
observational: to explain, predict, and maintain what is present in the sensory
stream under partial and uncertain evidence.

At the agentic level, the system additionally selects actions. These may be
chosen to complete a task, to acquire evidence that resolves uncertainty, or
to do both. The system may decide where to attend, which sensor or viewpoint
to query, which object to manipulate, or which hypothesis to test. An action
may succeed while revealing little, advance the task while improving the
perceptual estimate, or fail while exposing an important error in the current
interpretation. Active perception couples task-directed action to its perceptual consequences 
\citep{bajcsy2018revisiting,friston2021active}. V-JEPA 2 \citep{assran2025vjepa2} connects learned latent video dynamics to prediction and planning, while systems such as Gemini Robotics \citep{abdolmaleki2025geminirobotics15} connect visual–language and spatial reasoning to task planning, progress estimation, and motor action. Such systems may continually update their current state estimate and replan while acting.  

The levels form a progression. Passive temporal evidence
already contains the consequences of action and can reveal substantial world
structure from observation alone. Active sensing extends this capacity by
acquiring evidence targeted to both task completion and unresolved
hypotheses. A deeper loop arises when action outcomes and perceptual errors
alter not only the current state estimate, but durable memory, predictive dynamics,
representations, or eventually the system's ontology. The present paper
focuses primarily on the perceptual substrate required across these loops,
while treating cumulative learning through exploration and feedback as an
important but still comparatively underdeveloped extension.

\section{Why Multimodal Perception Is Still Unsolved}
\label{sec:why_perception_unsolved}

Perception itself remains unsolved even before action closes the loop. Current systems still lack a general formulation of multimodal perception as the construction of explicit, temporally coherent, uncertainty-aware open-world state from heterogeneous sensory streams.

Modern multimodal models can recognize objects, retrieve images from text, caption videos, answer visual questions, segment prompted regions, align audio with visual events, and synthesize plausible images or videos. These are major achievements, but they are not equivalent to solving perception. 

\begin{figure}[t]
    \centering
    \includegraphics[width=\linewidth]{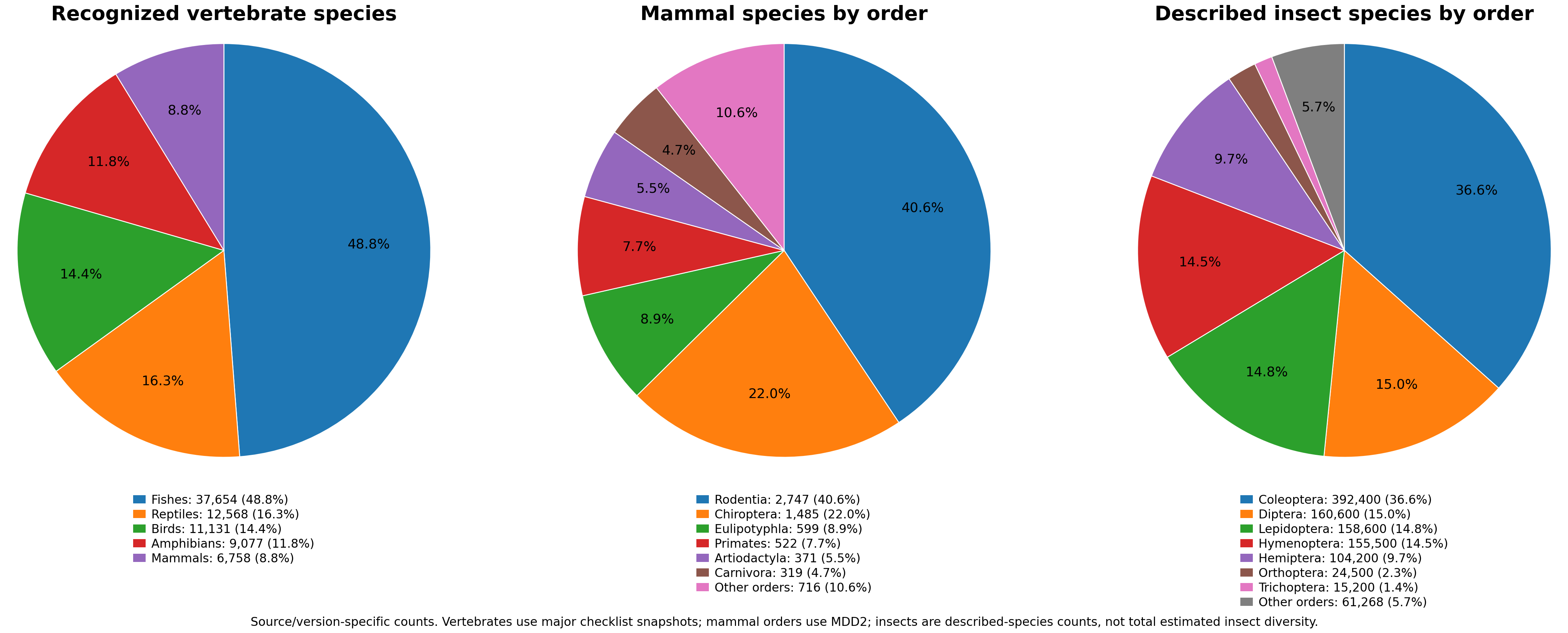}
    \vspace{-5mm}
    \caption{
    \textbf{The open-world category and statistics gap.}
    Biological diversity is highly uneven across taxonomic groups and remains only partially characterized at the level needed for perceptual modeling.
    The left panel shows a source-specific snapshot of reported vertebrate species across major vertebrate groups; the middle panel shows mammal species by major order using the Mammal Diversity Database MDD2 taxonomy; the right panel shows approximate described insect species by major order, not estimates of total extant insect diversity.
    Even within comparatively well-studied groups, taxonomic richness is highly uneven, and source checklists remain incomplete, aggregated at different levels, and subject to revision. 
    More importantly, species names alone do not provide the quantitative structure required for perception: morphology, motion, behavior, interaction, and multimodal sensory statistics are often missing.
    The figure illustrates why open-world multimodal perception cannot be reduced to fixed supervised label spaces.
    Counts are source- and version-dependent, based on major taxonomic checklists and compilations for fishes, birds, amphibians, reptiles, mammals, and insects \citep{eschmeyer2026catalog,avilist2025,amphibiaweb2026,reptiledatabase2026,burgin2025mdd2,zhang2013arthropoda}.
    }
    \label{fig:species_distribution_gap}
\end{figure}

\paragraph{The ontology and statistics gap.}
A major challenge is the open-endedness of the perceptual world. Current visual benchmarks often create the impression that perception can be reduced to a finite list of categories, attributes, and tasks. This impression is useful for benchmarking but misleading as a theory of perception. COCO contains 80 object categories over roughly 330K images~\citep{lin2014coco}; LVIS expands this vocabulary to 1,203 object categories and explicitly exposes the long-tailed nature of object categories~\citep{gupta2019lvis}; ImageNet-1K contains 1,000 classification categories, while the broader ImageNet resource spans over 14 million images and 21,841 synsets~\citep{deng2009imagenet,russakovsky2015imagenet}. Yet even these datasets represent only a thin slice of the physical and biological world (fig. \ref{fig:species_distribution_gap}). 

Biological diversity is highly long-tailed, and the quantitative structure needed for perception is often missing even for known categories. The gap concerns both category coverage and the statistics that make categories perceptually meaningful: morphology, motion, articulation, multimodal signatures, behavior, and interaction context. A taxonomic record may rest on one or a few specimens, diagnostic anatomy, a small set of photographs, or a historical illustration present in a museum collection. Perceptual learning, at least currently, requires variation across viewpoint, lighting, pose, behavior, life stage, and ecological context, together with structured annotations or other evidence concerning shape, motion, articulation, sound, and interaction. This is not practical and does not scale.

Biological diversity provides a well-defined illustration of a broader open-world problem. Man-made environments and artifacts exhibit a different form of combinatorial structure shaped by design, manufacturing, culture, regulation, and use. A category such as chair,'' door,'' vehicle,'' or tool'' is only the entry point to a richer perceptual model involving structure, state, function, and context. Recognizing an object as a door, for example, does not establish whether it swings or slides, how its mechanism operates, or whether it is currently obstructed. Category inventories, shape collections, and engineering descriptions capture aspects of this structure, but a perceptual system must infer which properties and states apply to the particular object encountered. Closed-vocabulary recognition restricts predictions to a predefined set of categories; open-vocabulary recognition broadens the concepts accessible through language without, by itself, establishing the geometry, dynamics, functional properties, or sensory statistics of the particular entities encountered. Perception must therefore go beyond mapping observations to names: it must infer persistent structure and changing state, represent uncertainty, and expand its model when new regularities appear.

\paragraph{The interface gap: linguistic reports are not full perceptual state.}
Many multimodal systems are evaluated through a linguistic interface. The model receives an image, video, audio clip, or interleaved context and produces text. This interface is powerful because language is flexible, compositional, and easy to evaluate at scale. However, it also risks confusing verbal description with perception. A fluent answer may reflect partial evidence, dataset priors, or linguistic plausibility without being supported by a grounded representation of the scene. This is one reason hallucination is particularly problematic in large vision-language models: the output can be linguistically coherent while failing to correspond to the actual sensory content \citep{li2023pope}. Language is essential for naming, abstraction, communication, instruction following, and semantic grounding. Within a perceptual system, it should serve as a conditioning, naming, abstraction, and explanation channel over perceptual state. In turn, geometry, correspondence, entities, events, uncertainty, and temporal persistence require representations appropriate to their underlying structure.

\paragraph{The multimodal evidence and binding gap.}
Contrastive image--text models, joint embedding spaces, and multimodal retrieval systems have shown that observations from different modalities can be placed in a common representational space. However, proximity or co-occurrence alone does not determine which components arise from, or refer to, the same entity or event. Nor does common tokenization make modalities interchangeable sources of evidence. Integrating their complementary constraints requires accounting for their differing uncertainty, resolution, and relation to the world.

Cross-modal binding assigns observations to sources and referents: which object produced a sound, which person uttered particular words, or which visual and tactile observations concern the same contact event. Existing methods address such associations in specific settings; the broader challenge is to resolve competing sources and ambiguous references when several entities interact, relevant sources are unobserved, or modalities provide conflicting evidence. This requires distinguishing signals that merely co-occur from those attributable to the same underlying source, while retaining uncertainty when the observations do not determine a unique assignment.

\paragraph{The temporal gap.}
Much of visual foundation modeling remains image-centric or short-context. Yet the world is not a collection of independent images. It is a temporally evolving process in which entities persist, deform, disappear, reappear, collide, interact, and cause sensory consequences across modalities. Static recognition cannot determine identity through occlusion, distinguish a transient texture from an object, infer a gait cycle, separate an action from its preparation, or identify the moment when one event causes another.

Temporal perception requires maintaining state. The system must infer what remains true when it is not visible, what has changed, what has remained identical, what is ambiguous, and what is likely to happen next. This is particularly demanding for articulated bodies, animals, tools, deformable materials, social interactions, and physical contact, where the relevant structure unfolds over time and is only partially visible in any single frame. A model that understands an event only by captioning the clip has not necessarily represented the event's geometry, dynamics, or causal organization.

Long context should not be conflated with memory. Contemporary multimodal
models can process increasingly long sequences, including long documents,
audio streams, and videos, and can retrieve or integrate evidence distributed
across those sequences
\citep{geminiteam2024gemini15,openai2024gpt4o}. This substantially extends
the temporal evidence available within one inference episode. A context
window, however, is still a bounded collection of supplied observations.
Perceptual memory additionally concerns what is retained, updated, compressed,
or forgotten when observations leave that window or when the system encounters
the same entities and environments again. It requires continuity of state
across processing episodes, not only access to a longer input sequence.

\paragraph{The Umwelt (environment-world) gap.}
Perception is not defined independently of the perceiver. In biology, the notion of an \emph{Umwelt} captures the idea that each organism inhabits a perceptual world shaped by its sensory apparatus, body, ecological niche, and possible forms of interaction \citep{uexkull1934stroll}. A bat, a dog, a person, an autonomous vehicle, and a humanoid robot may occupy the same physical environment, but they do not perceive the same world in the same way. Their relevant state variables, temporal scales, sensory channels, and possible interactions differ. This view is close to Gibson's account of the senses as active perceptual systems rather than passive input channels~\citep{gibson1966senses}: vision, hearing, touch, proprioception, and movement provide different ways of sampling structure relevant to an organism's body and possible action. 

This is not merely a biological curiosity, nor does it imply that artificial intelligence must simulate the perceptual worlds of all animals or systems. 
The methodological implication is that no single perceptual output format is equally appropriate for all perceivers. Human, robotic, assistive, and animal systems require different projections of the same environment. A general perceptual foundation model should support multiple projections of world state, together with calibration and fine-tuning. The model should represent objects, events, and relevant perceptual asymmetries: what other agents are likely to see, miss, intend, or misjudge. 
Such modeling need not amount to a full simulation of another Umwelt, but it is essential for communication, cooperation, and safe coexistence.

\paragraph{Operational criteria for a perceptual world state.} A multi-task model is not a world model merely because its tasks share a backbone. It becomes a perceptual world model only when its task-conditioned projections remain compatible with a persistent state that can be updated, queried and contradicted by new evidence. An operational definition of world state would satisfy several properties:

{\noindent \it Persistence.} The representation maintains continuity through missing observations, occlusion, re-entry, and successive processing windows, with each window updating an ongoing perceptual state.

{\noindent \it  Addressability.} Entities, surfaces, events or hypotheses can be referred to consistently across queries and time.

{\noindent \it  Cross-readout compatibility.} Depth, pose, segmentation, correspondence, event, sound, proprioception, and language readouts must be jointly compatible with one another.

{\noindent \it  Binding.} Evidence from different modalities and readouts is assigned to common underlying causes. Statistical alignment alone is insufficient.

{\noindent \it Revisability (evidence-driven revision).} New evidence updates, revises, or contradicts the existing state. Successive outputs remain linked through that state.

{\noindent \it  Epistemic structure.} The state represents point predictions together with ambiguity, alternative hypotheses, and abstention.

These criteria define perceptual state functionally rather than requiring it to
coincide with a single tensor, token set, or decoded output. In an implemented
system, state may be distributed across temporal latent fields, persistent
entity or query tokens, and explicit structured readouts, provided that these
representations are coupled, mutually constraining, and revised as parts of one
coherent hypothesis about the world. Retention of perceptual world state across separated encounters
introduces the additional requirement of durable episodic or semantic memory.

A foundational solution would assemble these components into a unified substrate. Such a system would infer, maintain, predict, and expose explicit world state from heterogeneous sensory streams; bind evidence across modalities; preserve identity through time; represent uncertainty and open-world structure; and produce outputs that can be inspected, queried, evaluated, and used for action. This is the standard against which the next generation of perceptual foundation models should be judged.

\section{The (Still) Missing GPT Moment of Perception}\label{sec:miss-gpt}

Language modeling became a foundation-model paradigm because representation,
objective, adaptation, and interface aligned unusually well. Text could be
represented as token sequences; large corpora supported self-supervised
pretraining; next-token prediction provided a scalable objective; and many
downstream tasks could be expressed through prompting, instruction tuning, or
fine-tuning in the same input--output space. Yet this alignment was only the
enabling condition for the GPT moment. Its defining transition was that a
single pretrained substrate could be elicited or adapted into dialogue,
instruction following, translation, coding, creation, and forms of reasoning
without engineering a separate model and supervised objective for every
capability. The success of GPT-like models combined scale and a reusable computational substrate with the emergence of broad capabilities that had not been explicitly decomposed into separately labeled downstream tasks \citep{brown2020language,wei2022emergent}.

Frontier systems have already moved decisively beyond a purely textual regime.
Recent GPTs were trained end-to-end across text, vision, and audio; Gemini was
designed and pretrained as a natively multimodal family, with later versions
operating over long contexts containing text, images, audio, and video; and
Claude supports substantial visual analysis while producing predominantly
linguistic outputs
\citep{openai2024gpt4o,geminiteam2023gemini,
geminiteam2024gemini15,anthropic2024claude3}. Modern predictive and embodied
systems already occupy part of this loop. V-JEPA~2 connects learned latent
video dynamics to prediction and planning, while systems such as Gemini
Robotics connect visual--language and spatial reasoning to task planning,
progress estimation, and motor action
\citep{assran2025vjepa2,abdolmaleki2025geminirobotics15}. The perceptual gap
cannot therefore be described simply as the absence of multimodal processing, joint multimodal training, sensory generation, or perception–action grounding.

At the level of computational substrate, however, the convergence remains
incomplete. Multimodal input is increasingly common, but multimodal output
remains fragmented across linguistic answers, generated sensory trajectories,
task-specific structured predictions, and motor actions. Multimodal assistants,
video-generative models, and embodied systems consequently emphasize different
interfaces and objectives: linguistic answers and tool use, dense
photorealistic trajectories, or task-directed action. Video-generative models
are optimized primarily for visual fidelity, temporal coherence, and
conditioning adherence \citep{google2025veo3,wan2025}; their internal
representations may encode geometry, motion, entities, sound--video synchrony,
and physical regularities without exposing them directly. Conversely, a
multimodal reasoning system may answer sophisticated questions about a video
while retaining mainly the abstractions required for the answer. These
boundaries may diminish, and proprietary systems may already combine the
corresponding capabilities more deeply than their public descriptions reveal.
What is not yet generally established is their organization around explicit
temporal perceptual state that remains persistent, queryable, revisable, and
compatible across linguistic, geometric, semantic, entity, event, sensory, and
generative readouts.

The missing transition is \emph{emergent perceptual generality}: a single pretrained substrate supporting a broad and expanding repertoire of perceptual capabilities beyond those made explicit during training. Current systems generally require dedicated supervision, output representations, losses, or adaptation for new perceptual questions and readouts. The enabling components are increasingly present, but their unification has yet to produce the perceptual analogue of the GPT transition.

\section{Current state of the art and its emerging opportunities}

Progress in visual perception has been considerable, but remains fragmented across domains and objectives. We briefly review advances in light of the desiderata outlined above and indicate how current trends motivate the broader hypotheses and directions proposed later in this paper.  Recent surveys on foundation models for vision~\citep{bommasani2021foundation,muhammad2024} highlight both the breadth of current progress and the persistent fragmentation of visual foundation models across semantic recognition, geometric estimation, and temporal understanding.

\noindent\textbf{Open-world semantics.}  Vision--language pretraining with CLIP~\citep{radford2021clip} and ALIGN~\citep{jia2021align} enables zero-shot transfer, while open-vocabulary detectors such as GLIP \citep{li2022glip} and Grounding DINO~\citep{ren2024groundingdino} extend this to detection. ODISE~\citep{xu2023odise} leverages diffusion features for panoptic segmentation, and SAM~2~\citep{kirillov2024sam2} offers class-agnostic segmentation and tracking. VideoCUPS further shows that depth, motion, and appearance cues can support temporally consistent panoptic pseudo-labels without human supervision~\citep{reich2026videocups}. OVTrack~\citep{li2023ovtrack} links names to tracks, providing entity-level persistence. Continual detectors such as CL-DETR~\citep{liu2023cldetr} address category growth while mitigating forgetting through distillation and exemplar replay. Despite these breakthroughs, current semantics are mostly flat: categories are represented primarily as label sets, while support for structured hierarchies, continual growth, and abstention is generally partial or addressed in isolation. Future systems could couple naming with physical and articulatory cues, grounding semantics jointly in text, geometry and dynamics.

\noindent\textbf{Dense and abstract 3D geometry.}  Pixel-wise geometry has advanced substantially.  MiDaS/DPT~\citep{ranftl2019towards,ranftl2021dpt}, ZoeDepth~\citep{bhat2023zoedepth}, and Depth-Anything~\citep{yang2024depthanything} demonstrate strong cross-dataset generalization for monocular depth. Diffusion-based DDVM extends this direction to monocular depth and optical flow while representing ambiguity through sampling~\citep{saxena2023ddvm}. Beyond depth, pixel-aligned reconstruction methods achieve detailed surface capture of people from images or video. PIFu~\citep{saito2019pifu}, ICON~\citep{xiu2022icon}, ARCH~\citep{huang2020arch}, and PHORHUM~\citep{alldieck2022phorhum} show that person-specific geometry can be recovered while generalizing across unseen people and images, while Garment Recovery uses learned shape and deformation priors to capture loose clothing~\citep{li2024garmentrecovery}. VGGT~\citep{wang2025vggt} predicts camera poses, depth maps, and dense correspondences in one inference pass, suggesting that unified 3D scene understanding is feasible. Dynamic-scene bundle adjustment provides a complementary route for estimating camera motion and producing temporally coherent dense reconstructions from casual video~\citep{chen2025backontrack}.  

Parametric models such as SMPL~\citep{loper2015smpl}, SMPL-X~\citep{pavlakos2019smplx}, and GHUM~\citep{xu2020ghum} offer strong priors on human articulation; ScoreHMR shows how diffusion priors over human parameters can guide recovery from single images, multiple views, and video~\citep{stathopoulos2024scorehmr}. Category-specific animal models such as 3D Safari~\citep{zuffi2019three} extend related ideas to animal pose, shape, and texture. Probabilistic directed distance fields provide a complementary implicit representation, coupling efficient differentiable depth rendering with flexible topology and a probabilistic treatment of surface and occlusion discontinuities~\citep{aumentadoarmstrong2022pddf}. Yet dense and abstract geometry remain siloed, with depth and normals detached from skeletons or implicit surfaces. Emerging generative 3D and 4D methods based on Gaussian splatting ~\citep{karnewar2024dreamgaussian,poole2024gaussianflow} point toward increasingly unified representations of geometry, appearance, and dynamics. A related line of work studies object-centric representations and learned latent dynamics, including slot-based models and predictive world models \citep{locatello2020slot,ha2018world,hafner2020dreamer,hafner2023mastering}. These approaches make entity structure more explicit than standard foundation models, but often remain limited in scale, modality breadth, or open-world transfer. 

\noindent\textbf{Temporal representation, prediction, and interactions.} Self-supervised video models such as VideoMAE~\citep{tong2022videomae}, together with architectures and objectives such as MaskFeat and TimeSformer~\citep{wei2022maskfeat,bertasius2021timesformer}, learn motion-sensitive representations across extended temporal windows. LV-MAE explicitly separates short-range encoding from long-range masked modeling to scale self-supervised representation learning to much longer videos~\citep{naiman2025lvmae}. Large-scale datasets such as AVA~\citep{gu2018ava}, AVA-Kinetics~\citep{li2020ava}, and Ego4D~\citep{grauman2022ego4d} have broadened evaluation beyond clip-level action recognition to temporal localization, anticipation, and egocentric interaction understanding. Grounded video captioning provides another bridge toward structured temporal semantics: GROVE grounds mentioned objects using temporally dense and consistent bounding boxes~\citep{kazakos2025groundedvideo}. Yet a system may recognize, localize, anticipate, or describe an interaction without explicitly representing its participating entities, their roles and relations, the state transitions that constitute the event, or uncertainty about its continuation. 

\noindent\textbf{Unified multimodal backbones.}  Foundation models have shown the value of shared representation spaces: CLIP~\citep{radford2021clip}, ALIGN~\citep{jia2021align}, PaLI-X~\citep{chen2023palix}, and Kosmos-2~\citep{peng2023kosmos2} for image--text; InternVideo2~\citep{wang2024internvideo2} and mPLUG-2~\citep{xu2023mplug2} for video--language. EgoM2P extends this direction to temporally tokenized egocentric RGB, depth, camera pose, and gaze, supporting both perception and conditional synthesis~\citep{li2025egom2p}. Flamingo \citep{alayrac2022flamingo} and VideoCoCa \citep{yan2022videococa} extend multimodal pretraining, while Sapiens~\citep{khirodkar2024sapiens} unifies multiple human-centric visual tasks within a shared backbone. Latent diffusion established a scalable approach to conditional visual synthesis in compressed representation spaces~\citep{rombach2022latentdiffusion}. Generative video models such as Sora \citep{openai2024sora}, Lumiere~\citep{bar2024lumiere}, \emph{Veo}~\citep{google2025veo3}, \emph{Wan}~\citep{wan2025}, and Seedance~\citep{gao2025seedance} produce temporally coherent sequences but are evaluated primarily for visual fidelity, temporal coherence, and conditioning adherence rather than explicit structured state. Instruction-tuned models such as LLaVA-Video \citep{zhang2024llavavideo} and Video-LLaMA~\citep{zhang2023videollama} bring language conditioning to temporal understanding, yet their outputs remain predominantly verbal, with limited geometric or semantic overlays. 

Beyond image--text alignment, multimodal tokenization and unified embeddings continue to mature.  ImageBind ~\citep{girdhar2023imagebind} aligns six modalities (image, text, audio, depth, thermal, IMU) without requiring paired data for every modality combination, using image-paired data as a common bridge. Audio–visual representation learning provides a more direct precedent for joint temporal modalities: CAV-MAE combines masked autoencoding and contrastive learning to learn coordinated audio–visual representations \citep{gong2023cavmae}, while SoundStream and EnCodec provide learned tokenization and compression mechanisms for audio ~\citep{zeghidour2021soundstream,defossez2022encodec}. Speech foundation models provide complementary examples. HuBERT learns transferable speech representations through self-supervised masked prediction of hidden units ~\citep{hsu2021hubert}, while Whisper obtains robust multilingual and multitask speech recognition from large-scale weak supervision ~\citep{radford2022whisper}.

4M and 4M-21~\citep{zhao2023fourm,bachmann2024fourm21} show that semantic, geometric, textual, feature-level, and pose-like modalities can be cast as discrete token prediction within an any-to-any masked modeling framework. FlexTok and VideoFlexTok further suggest that visual and video tokenization need not remain a fixed dense grid: variable-length, ordered, coarse-to-fine token sequences can allocate capacity according to scene complexity and expose abstract information such as semantics and motion before fine detail \citep{bachmann2025flextok,atanov2026videoflextok}. These works provide important precedents for token-level unification, while leaving temporal persistence, entity state, and structured readout as open architectural questions. 

\citet{acuaviva2025generation} provide closely related evidence from video diffusion models, showing that pretrained video generators encode reusable visual priors through few-shot adaptation to diverse tasks using input–output video transitions and LoRA fine-tuning. This supports the broader view that synthesis-oriented temporal pretraining can provide transferable structure for perception, complementing the THFM and GenCeption line of work \citep{wang2026thfm,wang2026eccv}. The generative route also has characteristic failure modes. Analyses of convolutional WGANs show how realistic generation can arise from matching patch distributions rather than whole-image distributions, underscoring the role of architectural inductive bias~\citep{elnekave2025wgan}. A video model may produce plausible continuations without representing metric geometry, causal mechanism, or persistent identity in a reliable way. Photorealism can mask structural error, and temporal smoothness can be achieved through shortcuts that do not support intervention or calibrated uncertainty. A relevant inference question is whether the backbone supports stable, inspectable, and cross-validated perceptual structure.

\noindent\textbf{Reliability and robustness.}  Despite significant progress, reliability remains a critical bottleneck.  Systems often fail silently under distribution shift, producing plausible but incorrect outputs.  Open-set and OOD detection methods ~\citep{bendale2016towards,hendrycks2017baseline}, together with robustness benchmarks such as ImageNet-O \citep{hendrycks2021natural} and ObjectNet \citep{barbu2019objectnet} expose failures under distribution shift and controlled variation. Recent research studies predictive uncertainty under dataset shift \citep{ovadia2019can}, OOD detection \citep{fort2021exploring}, and selective prediction with explicit abstention \citep{geifman2019selectivenet}, though large multimodal models still lack widely validated, scalable uncertainty estimates.

\noindent\textbf{Understanding emergence and interpretability.}  Interpreting how models represent structure has become as important as measuring task performance. Saliency-based methods~\citep{zeiler2014visualizing,selvaraju2017gradcam,sundararajan2017ig} and representation-level metrics (SVCCA~\citep{raghu2017svcca}, CKA~\citep{kornblith2019cka}, RSA~\citep{kriegeskorte2008rsa}) provide tools for probing and comparing internal representations. Self-supervised ViTs such as DINO ~\citep{caron2021dino} exhibit spontaneous object segmentation; Slot Attention provides an explicit mechanism for object-centric grouping ~\citep{locatello2020slot}, while recent interpretability work has identified semantically meaningful internal representations in diffusion models ~\citep{chefer2024hiddenlanguage}. Scaling studies~\citep{wei2022emergent,olsson2022superposition} reveal phase transitions where new abilities appear, offering tools to probe emergence in perception.  Recent analyses of pretrained video models and generators provide more direct evidence of reusable perceptual structure, visual transfer, and structured perceptual readouts \citep{acuaviva2025generation,wiedemer2025video,wang2026thfm,wang2026eccv}. Understanding these emergent effects will be key to diagnosing and guiding open-world perceptual intelligence.

\noindent\textbf{State estimation and mapping.} The requirement to maintain a revisable belief state is consistent with Bayesian filtering, simultaneous localization and mapping, and planning under partial observability \citep{thrun2005probabilistic,durrantwhyte2006slam,kaelbling1998planning}. These provide explicit principles for updating state under noisy sequential evidence, generally within a specified state and observation structure even when the state itself, data association, or map extent remains uncertain. Learned world models extend this formulation by acquiring predictive latent state and dynamics from experience \citep{ha2018world,hafner2020dreamer,hafner2023mastering,bruce2024genie}. The present proposal further asks whether temporal foundation models can support multimodal, open-world perceptual state whose entities, correspondences, structured readouts, and event organization are only partly specified in advance. They are proposed as scalable representational substrates for an enlarged state-estimation problem, not as replacements for its probabilistic foundations.

\noindent\textbf{World models.} This research intersects with the broader pursuit of world models~\citep{ha2018world,lecun2022path,hafner2020dreamer,hafner2023mastering,bruce2024genie}, which couple perception, prediction, and sometimes control. World models for action require states that are compact enough for prediction and planning, but grounded enough to support reliable intervention in the external world.  A compact latent dynamics model may be useful for prediction, imagination, or policy learning, but an acting system must also infer which entities exist, where they are, how they move, what they afford, what is uncertain, and how sensory evidence should revise the current state. In practice, this requires perceptual interfaces---for example readouts, query mechanisms, or structured heads---that expose object state, geometry, correspondence, contact, events, affordances, and uncertainty from the learned world state. Supporting structured perception natively within a coherent perceptual training program may preserve coherence, efficiency, and generalization while allowing subsequent refinement through use, exploration, and action. Under this latter view, structured perception is not separate from world modeling: it is one of the ways in which a world model becomes usable for reasoning, communication, diagnosis, planning, and action, while also providing a substrate for continued refinement as new objects, situations, interactions, and failure modes are encountered.

\noindent\textbf{Visual representation and policy learning.} A complementary research line couples visual representation learning directly to control. Deep visuomotor policies demonstrated that perception and motor control can be trained jointly from images to robot actions, reducing reliance on separately engineered state estimators \citep{levine2016endtoend}. Related work learned compact spatial representations from camera observations and used them as task-relevant state for reinforcement-learning-based control \citep{finn2016deep}. Model-based reinforcement learning extends this connection by learning latent visual representations and predictive dynamics from interaction, then optimizing policies within the resulting model \citep{hafner2020dreamer,seo2023masked}. These systems show that action can shape which visual structure becomes useful and that learned perceptual representations can support closed-loop control. The emphasis of the present paper is complementary: to develop a broader, explicit perceptual substrate that can support multiple tasks, agents, and readouts while remaining available to policy learning and closed-loop control.

\section{Guiding Hypotheses and Conceptual Architecture}
\label{sec:framework}

Generalist open-world perceptual intelligence treats perception as inference of structured world state constrained by evidence. This state includes geometry, entities, dynamics, relations, and uncertainty, represented at two coupled levels: an internal, latent, temporal token space in which evidence is integrated, retained, and revised, and an explicit perceptual level exposing inferred structure through dense and sparse readouts. Their coupling makes latent structure addressable and allows individual readouts to remain compatible projections of a common state.

\subsection{Hypotheses}{\label{subsec:hypotheses}

The view of perception as inference of structured open-world state motivates four interdependent, empirically testable hypotheses about perceptual properties that may emerge within the proposed class of temporal models.

\textbf{H1. Perceptual structure can emerge within generative backbones.}
Large generative models---diffusion, autoregressive, and masked-prediction transformers---often exhibit internal signals that correlate with perceptual structure. 
Attention patterns may align with object boundaries; latent features may correlate with depth, motion, or semantics; and temporal models may preserve information useful for identity and scene continuity. 

A plausible hypothesis is that, given sufficient data diversity and temporal coherence, generative objectives favor internal coordinates that preserve some of the factors needed to predict sensory evolution—including geometry, appearance, motion, persistence, and interaction. These coordinates need not be fully disentangled or causally identified. The substantive question is whether structured adaptation, for example through supervised fine-tuning, can additionally promote disentanglement and expose, control, and recombine them in ways that support reliable perceptual inference.

Testing this hypothesis requires diagnostic tools that measure \emph{structural emergence}: for instance, assessing how consistently a model’s latent variables track entities, parts, or surfaces across time. 
If confirmed, this would imply that rich generative training can provide much of the substrate for grounded perception, while supervised or self-supervised adaptation may mainly expose, align, and validate the relevant structured variables.

\paragraph{H2. Temporal coherence and multimodal correlation induce entity formation.}
In dynamic environments, persistence is one of the strongest cues for candidate objecthood. When regions of an image move coherently across frames, they signal a single entity. 
When a sound stays synchronized with that region—footsteps with a moving person, engine noise with a car—the evidence becomes multimodal. 
Our hypothesis is that such temporal and cross-modal coherence can provide a powerful training signal for entity discovery, even when entity labels are absent or incomplete.  

Generative transformers trained to model temporally coherent video, and in some cases audio--visual streams, must preserve information useful for temporal consistency. 
Their internal states may contain proto-object-like signals: features or token patterns that preserve information about entities through occlusion, deformation, or viewpoint change. 
These signals may support persistence, continuity, and identity.

The complexity of object discovery parallels the tension between \emph{pattern completion} and \emph{pattern separation} in memory trace formation \citep{mcnaughton1987hippocampal,oreilly1994hippocampal,yassa2011pattern}. 
A perceptual system must learn to complete an entity across nuisance variation while also separating genuinely different causes or structures when the evidence requires it.
Entity formation therefore cannot be reduced to temporal smoothness. 
It requires a representation flexible enough to absorb changes in appearance, pose, and visibility, yet sharp enough to split identities, parts, or events when they reflect distinct underlying causes.

Similar principles have long been pursued in motion-based grouping. Psychophysical and developmental work emphasized principles such as common fate, cohesion, continuity, and contact in object perception \citep{johansson1973biological,spelke1990principles}. Computational vision developed related ideas through structure-from-motion, optical flow, graph partitioning, long-term point trajectories, and spatio-temporal clustering \citep{ullman1979structure,black1996robust,shi2000normalized,brox2010object,ochs2012higher,hartley2004multiple,szeliski2022computer,torralba2024foundations}. These approaches established that motion is a powerful cue for discovering object boundaries and persistent structure. However, they also exposed the difficulty of balancing heterogeneous evidence: appearance, flow, geometry, rigidity, articulation, contact, occlusion, and category priors often have to be hand-combined or tuned in complex pipelines. This has not yet yielded a robust foundation-level mechanism for automatically discovering persistent entities across the full range of open-world temporal variation.

Our evidence suggests that temporal perception backbones of the kind proposed here can turn temporal predictability into structure discovery. Alternative paths may explore attention-based grouping, dynamic slot allocation, or explicit memory mechanisms. 
In either case, the hypothesis is that temporal regularities provide a mechanism for entity formation, linking self-supervised prediction to persistent structure in the physical world.

\paragraph{H3. Articulation and correspondence priors are partially universal and transferable.}
Across species, objects, and mechanisms, the physical world repeatedly expresses constrained motion and persistent surface structure: articulated motion through joint-like or deformable relations, locally rigid parts, surface continuity, bilateral or axial organization, and contact constraints.

Our hypothesis is that these regularities are sufficiently common to support reusable priors expressed through an articulation and correspondence manifold: a latent space of deformation laws, surface relations, and motion constraints that describes how parts, surfaces, and wholes remain related through time. When a model trained on human motion learns contact, surface persistence, and part-relative motion, some of these priors can generalize to animals, robots, tools, and other complex systems. 
The transferable structure is a family of constraints: articulated parts often remain related through limited degrees of freedom, surfaces persist through motion, and contact restricts possible dynamics. 
Different categories will require different correspondence domains, but the underlying idea of stable structure through time can be reused.

Dense correspondence may be especially important in this transfer, but the topological and semantic claims should be separated. Many biological and articulated objects, when considered at a coarse external-surface resolution, can be approximated by low-genus surfaces, often close to a genus-zero abstraction. This is a perceptual approximation, not an anatomical or physical claim. At higher levels of detail, digestive tracts, clothing layers, and manufactured mechanisms can introduce higher-genus topology, discontinuities, or chart boundaries. Nevertheless, the externally visible surface can often be parameterized by a canonical domain, learned surface embedding, deformable template, or set of local charts. Even then, it does \emph{not} mean that all such objects should share the same correspondence map, or that the same semantic part must occupy the same region of a universal sphere, for example. Broadly different categories may require their own internally consistent correspondence domains: humans, primates, or humanoid robots may share one family of surface coordinates, birds another, fish another, and tools or mechanisms yet another. 
Related categories may admit reusable correspondence priors once expressed through suitably generalized local or category-specific domains.

\begin{figure*}[t]
\centering
\begin{minipage}[t]{0.54\textwidth}
\centering
\includegraphics[width=\linewidth]{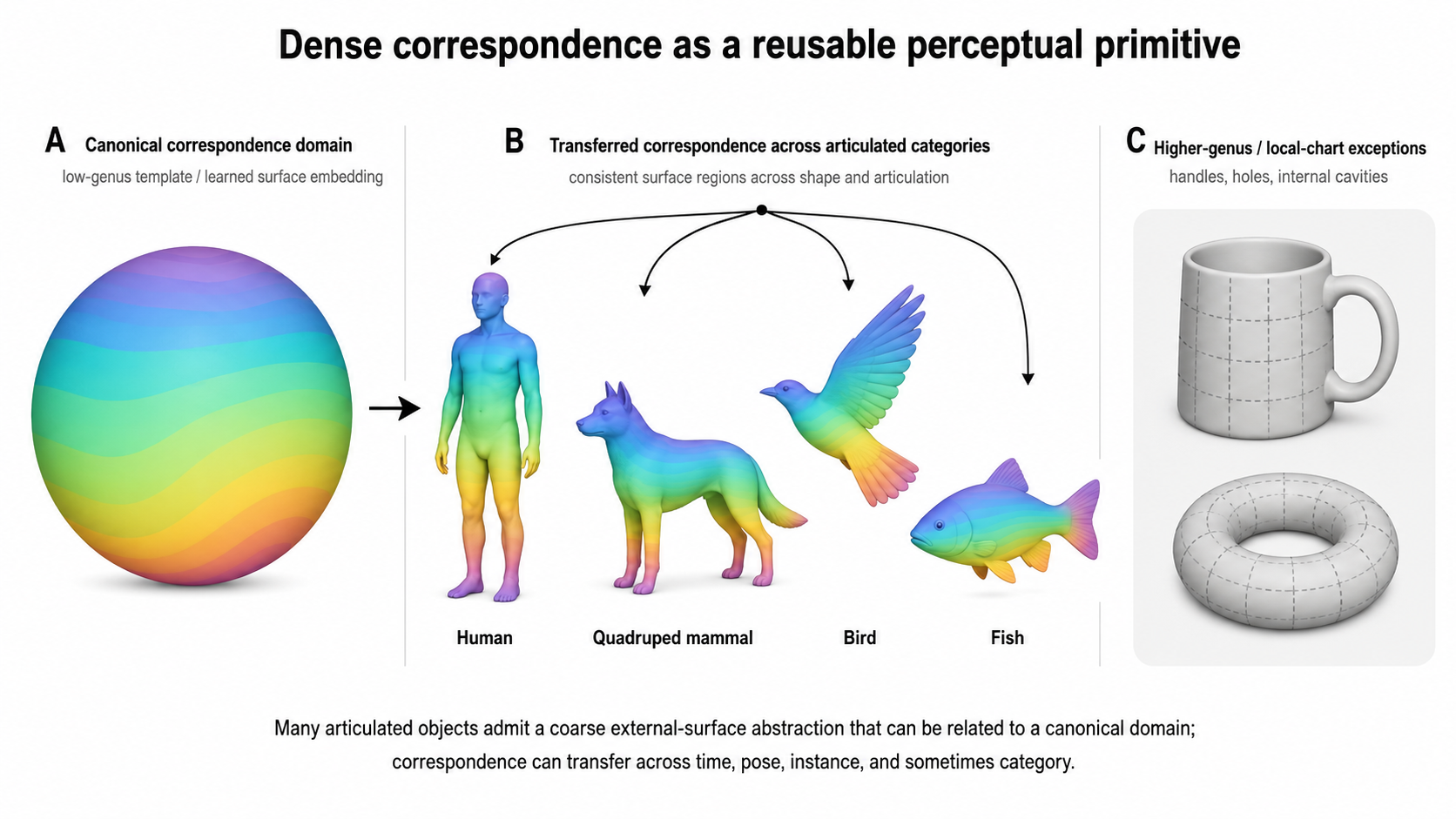}
\end{minipage}
\hfill
\begin{minipage}[t]{0.44\textwidth}
\centering
\includegraphics[width=\linewidth]{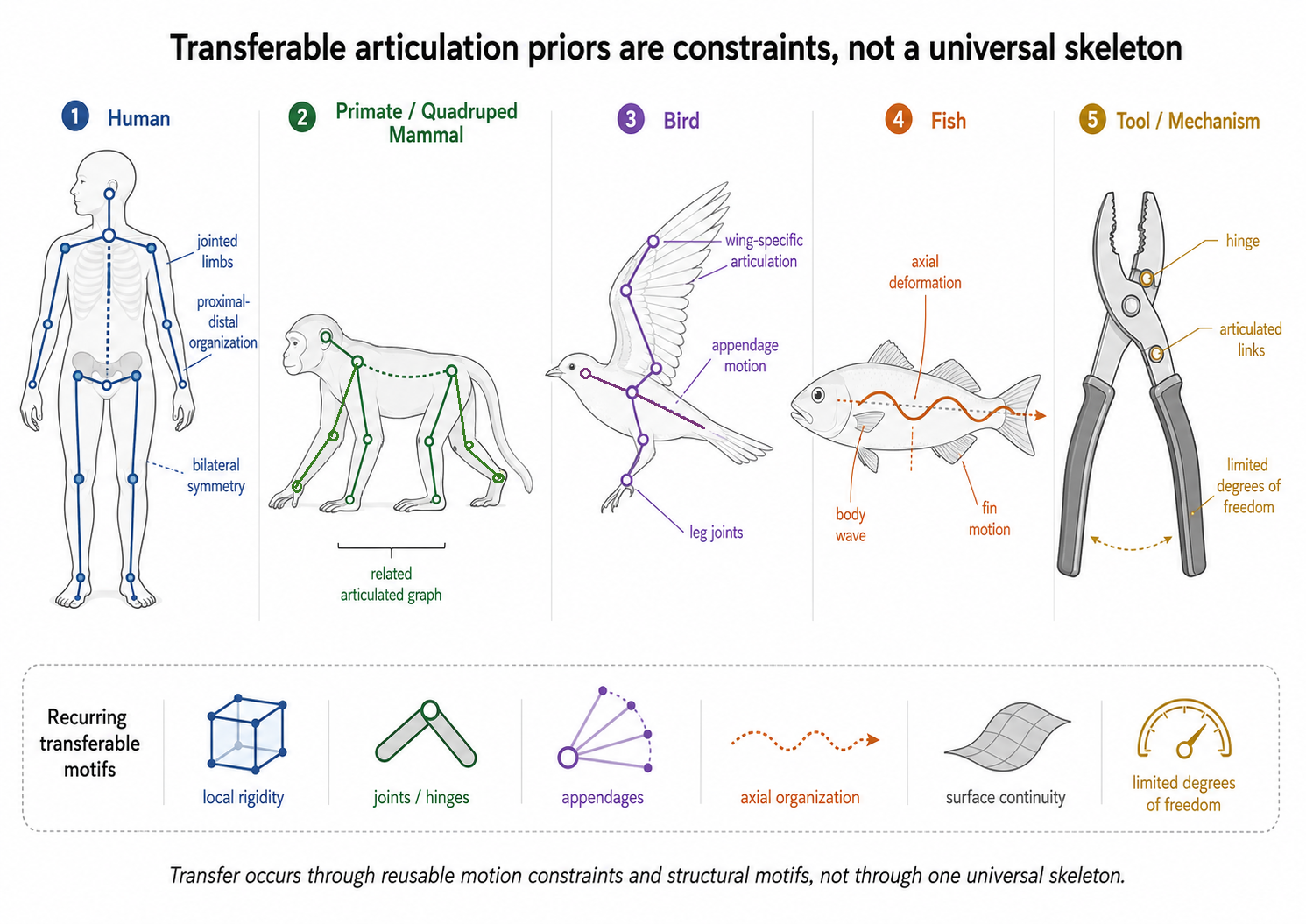}
\end{minipage}
\vspace{-0.5em}
\caption{
\textbf{Conceptual illustration of correspondence and articulation as partially transferable perceptual structure.}
Left: dense correspondence can provide a reusable coordinate structure across time, pose, viewpoint, instance variation, and sometimes related categories, while higher-genus objects may require cuts, patches, or local charts.
Right: articulation transfer should be understood as the reuse of motion constraints and structural motifs---such as local rigidity, joints, axial deformation, and limited degrees of freedom---rather than as the instantiation of one universal skeleton across all categories.
}
\label{fig:correspondence_articulation_priors}
\end{figure*}

\paragraph{H4. Correlated structured outputs align latent representations and expose reliability signals.} 
When a model is trained to predict multiple correlated quantities---segmentation, depth, surface normals, optical flow, correspondence, pose, or camera state---it is encouraged to represent shared explanatory structure across outputs. 
Shared latent capacity, task prompting, and consistency among outputs can align internal representations, although they do not guarantee a coherent world model by themselves. 
The empirical question is when multi-output training produces positive transfer, when it merely shares computation, and when it creates interference.

Our hypothesis is that multi-output generative conditioning can organize the latent representation itself, with effects extending beyond individual task accuracy. By learning to produce several complementary outputs from a shared code, the model is encouraged to acquire a common representation of space and semantics. 
Depth and segmentation constrain each other; optical flow informs articulation; sound and text refine spatial and temporal boundaries. 
Agreement and disagreement among modalities and structured outputs can also expose reliability signals. Such agreement is not itself calibrated confidence: correlated readouts may share a bias and remain consistently wrong. Its value is diagnostic, providing signals whose relation to actual error can subsequently be measured and calibrated.  Empirically, this can be studied through probing and intervention—measuring how perturbations in latent geometry affect semantic decoding or vice versa. 
If these dependencies prove stable and interpretable, they could support systems that detect likely failures through cross-modal and cross-readout inconsistency, with confidence calibrated separately against observed error.

\paragraph{Joint contribution and falsifiability.} We advance these four hypotheses because they already have varying degrees of empirical support in our prior work, reinforced by complementary findings in the broader literature—for example, diffusion models trained to output segmentation maps, articulation transfer in animal pose estimation, emergent tracking in video transformers, or multi-task synergies in dense prediction. They are not claimed to be jointly sufficient to satisfy all the perceptual requirements identified in \S\ref{sec:why_perception_unsolved}; a complete solution may require additional hypotheses, principles, or mechanisms. To our knowledge, no prior work has jointly probed these four hypotheses within a model for open-world temporal perception. Their systematic unification within one framework, where they are made explicit, offers the opportunity to instantiate concrete models and then test or falsify them through rigorous evaluation. This cycle of assumption, construction, and critical validation embodies the scientific method: it ensures that progress is measured not only by building larger models, but by understanding which principles genuinely enable open-world perception.

The hypotheses are also falsifiable. 
They would be weakened if temporal generative pretraining failed to improve structured readouts over non-generative or frame-based baselines; if apparent open-world transfer disappeared under controlled evaluation; if multi-output training improved task scores without improving cross-output consistency; or if latent probes revealed no stable entity, correspondence, or temporal structure beyond what is imposed by supervised heads.

\subsection{Conditional Generative Perception Architecture}
\label{subsec:architecture}

Large-scale video generators such as \emph{Veo}~\citep{google2025veo3} and \emph{Wan}~\citep{wan2025} demonstrate that photorealistic and temporally coherent video can be synthesized directly from text prompts. Wan makes one current implementation pattern explicit: video is encoded into spatio-temporal latents, a diffusion transformer operates on these latents under text and other conditioning signals, and a decoder maps the resulting latents back to RGB video. Thus, the central generative computation occurs primarily in latent space, while reconstruction, post-training, fine-tuning, preference optimization, and evaluation may remain tied to pixel-space video quality.

\paragraph{From synthesis to perception.}
We retain this pretrained backbone (latent video tokenizer, diffusion transformer, RGB decoder) but extend its purpose. Besides generating RGB conditioned on text, the model also learns to generate \emph{structured perceptual fields} from \emph{observed 
video} (and optionally audio): semantic masks, depth, surface normals, dense correspondences, and 2D/3D articulated poses. 
In perception mode, the usual text-conditioning interface is supplemented, or partly replaced, by observed video and audio conditioning tokens, allowing the diffusion transformer to produce temporally consistent and geometrically grounded structured outputs aligned with the input stream.

\begin{figure*}[t]
\centering
\includegraphics[width=\textwidth]{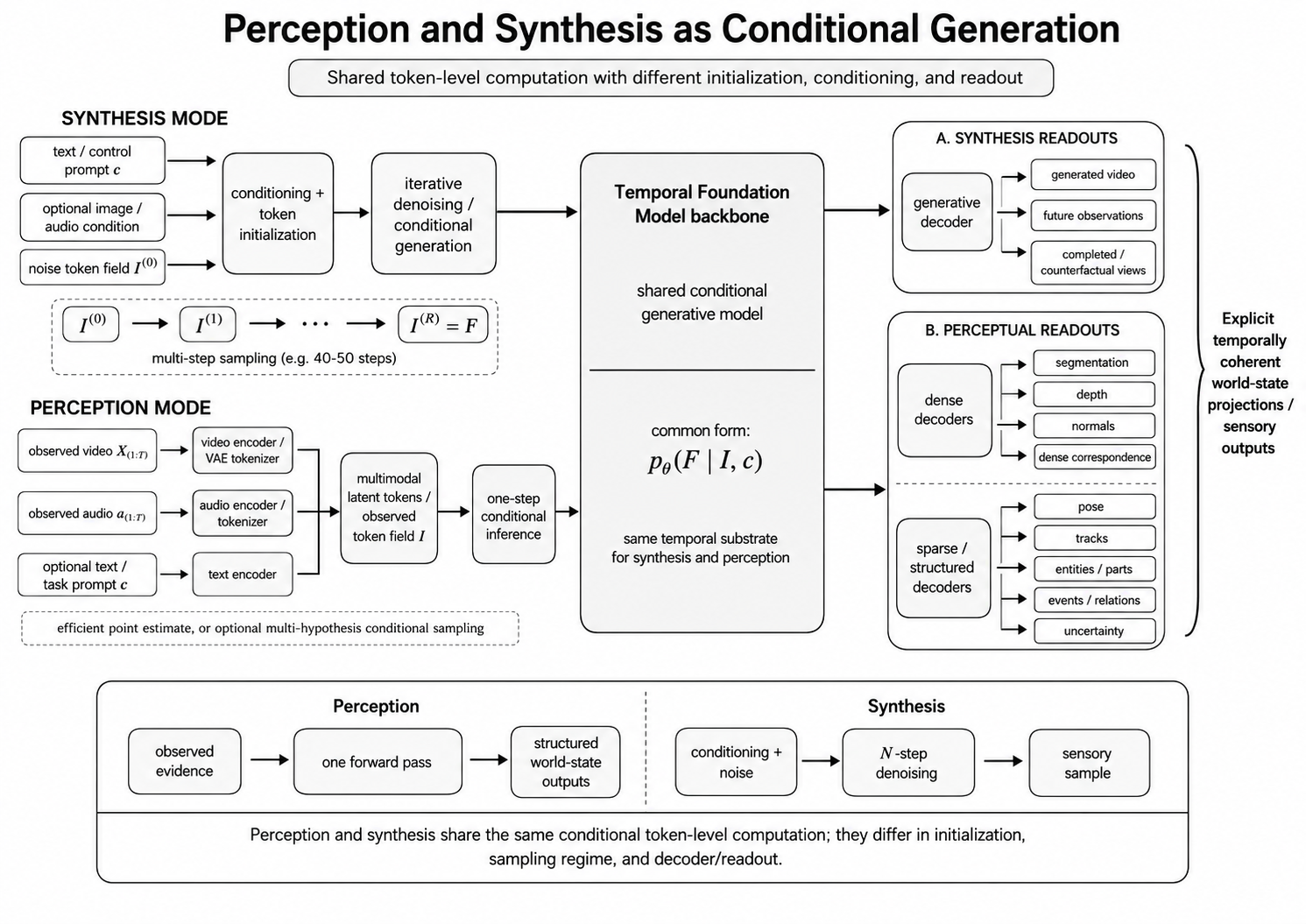}
\caption{
\textbf{Perception and synthesis as conditional generation.} A shared Temporal Foundation Model backbone can support both synthesis and perception as conditional generation over latent token fields. In synthesis mode, conditioning signals such as text, control inputs, or other modalities are combined with an initial noisy token field, and the model produces sensory outputs through iterative denoising and generative decoding. In perception mode, observed video, audio, and optional task prompts are encoded into an observed token field, and the same backbone architecture and synthesis-pretrained substrate produce structured world-state projections through efficient one-step conditional inference or, when needed, multi-step conditional sampling. Dense structured readouts include segmentation, depth, normals, and dense correspondence; sparse or entity-level readouts can include pose, tracks, entities, parts, events, or relations.  The latent field, together with any persistent entity or query tokens, acts as an internal carrier of perceptual state; the decoded outputs are explicit task-conditioned projections. Neither a single latent output field nor a single decoded readout is equated with the complete world state. The common conditional form is $p_{\boldsymbol{\theta}}(\mathbf{F} \mid \mathbf{I}, \mathbf{c})$, where $\mathbf{I}$ denotes the initial token field: noise in synthesis and observed sensory tokens in perception. Synthesis and perception differ mainly in initialization, sampling regime, and decoder/readout, while sharing the same temporally conditioned latent substrate. Structured perceptual layers may therefore operate in either direction: as
outputs inferred from sensory evidence or as controls supplied to synthesis.
} \label{fig:perception_synthesis_conditional_generation}
\end{figure*}

\paragraph{A common conditional token-field formulation.}
Dense perception and synthesis can both be formulated as temporally conditioned generation over latent token fields, while differing in initialization, parameter specialization, inference regime, and readout. Let $\mathbf{X}_{1\:T}$ denote an input video with $T$ frames, height $H$,
width $W$, and three color channels:
\begin{equation}
\mathbf{X}_{1:T}
\in
\mathbb{R}^{T \times H \times W \times 3}.
\end{equation}

Let $\mathcal{E}_{\boldsymbol{\phi}}$ denote the full temporal tokenization
map, comprising a temporal encoder or VAE tokenizer and any subsequent patch
embedding, with parameters $\boldsymbol{\phi}$. It maps the video to a
spatio-temporal latent token field
\begin{equation}\label{eq:2}
\mathbf{Z}
=
\mathcal{E}_{\boldsymbol{\phi}}(\mathbf{X}_{1:T})
=
[\mathbf{z}_1,\ldots,\mathbf{z}_K],
\qquad
\mathbf{Z} \in \mathbb{R}^{d \times K},
\qquad
\mathbf{z}_k \in \mathbb{R}^{d}.
\end{equation}
Here $\mathbf{Z}$ is the full token field, $\mathbf{z}_k$ is the $k$-th
column token, $K$ is the number of tokens passed to the temporal transformer,
and $d$ is the token dimension. If the VAE tokenizer produces a latent grid
of size $T' \times H' \times W'$ and the subsequent patch embedding uses a
patch size $P_t \times P_h \times P_w$ in the latent grid, then
\begin{equation}
K =
\frac{T'}{P_t}
\frac{H'}{P_h}
\frac{W'}{P_w}.
\end{equation}

For example, a video with $T=81$ frames and spatial size $H \times W$ may still yield thousands to tens of thousands of latent tokens after temporal and spatial compression, before further pooling, sparsification, or task-specific readout. The exact values of $K$, $T'$, $H'$, $W'$, and $d$ are architecture-dependent. The important point is that both synthesis and dense perception operate over structured temporal token fields rather than isolated frames.\footnote{As a concrete scale reference, the released Wan2.1 implementation uses a video VAE with
temporal--spatial compression $4 \times 8 \times 8$, 16 latent channels,
and a DiT patch size $1 \times 2 \times 2$ in $(T,H,W)$ order. Because
the causal VAE maps $T$ frames to $(T-1)/4+1$ latent time steps, an
81-frame video yields 21 latent steps. At resolution $832 \times 480$
(width $\times$ height), the VAE tensor therefore has shape
$16 \times 21 \times 60 \times 104$ in $(C,T,H,W)$ order, containing
$16 \cdot 21 \cdot 60 \cdot 104 = 2{,}096{,}640$ latent scalar values.
Patching produces a grid of
$21 \times 30 \times 52 = 32{,}760$ transformer tokens; each patch
contains $16 \cdot 1 \cdot 2 \cdot 2 = 64$ latent scalars before
projection. At resolution $1{,}280 \times 720$, the corresponding VAE
tensor has shape $16 \times 21 \times 90 \times 160$, and patching gives
$21 \times 45 \times 80 = 75{,}600$ tokens. In the Wan2.1 T2V-14B model,
each token is projected to transformer width $5{,}120$. Default sampling
uses 50 steps for text-to-video and 40 for image-to-video, so synthesis
repeatedly applies the temporal backbone to roughly
$3.3 \times 10^4$--$7.6 \times 10^4$ tokens per step, depending on
resolution.}

In video synthesis, let $\mathbf{I}^{(0)} \in \mathbb{R}^{d \times K}$ denote the initial latent token field from which sampling starts. In diffusion-based synthesis, this initial field is usually sampled noise:
\begin{equation}
\mathbf{I}^{(0)}
=
\left[\mathbf{i}_1^{(0)},\ldots,\mathbf{i}_K^{(0)}\right],
\qquad
\mathbf{I}^{(0)}
\sim
\mathcal{N}\!\left(\mathbf{0},\mathrm{I}_{dK}\right),
\qquad
\mathbf{i}_k^{(0)}\in\mathbb{R}^{d}.
\end{equation}
Here, $\mathbf{I}^{(0)} \in \mathbb{R}^{d \times K}$ is the initial
latent token field, $\mathbf{i}_k^{(0)} \in \mathbb{R}^{d}$ is its
$k$-th column token, and
$\mathrm{I}_{dK} \in \mathbb{R}^{dK \times dK}$ denotes the
identity covariance matrix.

Let $\mathbf{c}$ denote the conditioning signal, such as text, audio, an image, a motion cue, or a task specification. Let $\boldsymbol{\theta}$ denote the parameters of the temporal diffusion transformer backbone. The synthesis model defines a conditional distribution over an output latent token field $\mathbf{F} \in \mathbb{R}^{d \times K}$:
\begin{equation}
    p_{\boldsymbol{\theta}}
    \left(
        \mathbf{F}
        \mid
        \mathbf{I}^{(0)}, \mathbf{c}
    \right),
    \qquad
    \mathbf{F}
    =
    [\mathbf{f}_1,\ldots,\mathbf{f}_K],
    \qquad
    \mathbf{f}_k \in \mathbb{R}^{d}.
    \label{eq:synthesis_token_distribution}
\end{equation}
Here $\mathbf{F}$ is the generated latent token field, and $\mathbf{f}_k$ is its $k$-th column token. In diffusion sampling, $\mathbf{F}$ is obtained after $R$ reverse denoising steps, where $R$ denotes the number of sampling steps:
\begin{equation}
    \mathbf{I}^{(0)}
    \rightarrow
    \mathbf{I}^{(1)}
    \rightarrow
    \cdots
    \rightarrow
    \mathbf{I}^{(R)}
    =
    \mathbf{F}.
\end{equation}
The intermediate fields $\mathbf{I}^{(r)} \in \mathbb{R}^{d \times K}$, for $r=1,\ldots,R$, are denoised latent token fields. In contemporary video diffusion systems, $R$ is often on the order of a few tens of steps; for example, Wan2.1 uses about $40$--$50$ sampling steps depending on the image-to-video or text-to-video setting. Thus, synthesis repeatedly applies the temporal backbone to the same large token field across the denoising trajectory. Let $\mathcal{D}_{\boldsymbol{\psi}}$ denote a decoder with parameters $\boldsymbol{\psi}$. The decoder maps the generated token field back to an ambient output:
\begin{equation}
    \hat{\mathbf{X}}_{1:T}
    =
    \mathcal{D}_{\boldsymbol{\psi}}(\mathbf{F}).
\end{equation}
For synthesis, $\hat{\mathbf{X}}_{1:T}$ may be an RGB video, an audio stream, or another generated sensory modality.

For perception, the same token-field architecture and synthesis-pretrained substrate can be reused, while the initialization, inference procedure, and interpretation change. The input token field is no longer sampled noise. Instead, for notational alignment with the synthesis formulation, we relabel the encoded field from (\ref{eq:2}) as $\mathbf{I}:=\mathbf{Z}$ and its columns as
$\mathbf{i}_k:=\mathbf{z}_k$:

\begin{equation}
    \mathbf{I}
    =
    \mathcal{E}_{\boldsymbol{\phi}}(\mathbf{X}_{1:T})
    =
    [\mathbf{i}_1,\ldots,\mathbf{i}_K],
    \qquad
    \mathbf{I} \in \mathbb{R}^{d \times K},
    \qquad
    \mathbf{i}_k \in \mathbb{R}^{d}.
\end{equation}
Here $\mathbf{I}$ denotes the observed input token field, and $\mathbf{i}_k$ is its $k$-th column token. Depending on the instantiation, the observed visual field may be combined with tokens representing audio, language, haptic or proprioceptive inputs, task specifications, or structured queries. The same symbol $\mathbf{c}$ now denotes the perceptual conditioning signal: it specifies the requested readout, such as segmentation, depth, surface normals, dense correspondence, pose, tracks, entities, events, or uncertainty. The model estimates a conditional distribution over output perceptual tokens:
\begin{equation}
    p_{\boldsymbol{\theta}}
    \left(
        \mathbf{F}
        \mid
        \mathbf{I}, \mathbf{c}
    \right).
    \label{eq:perception_token_distribution}
\end{equation}
For perception, the organization of $\mathbf{F}$ is task- and architecture-dependent: it may retain a dense token grid, use a compact set of query or entity tokens, or combine the two. The output token field $\mathbf{F}$ is decoded not necessarily as RGB, but as structured perceptual variables. Let $\mathcal{D}^{(\mathbf{c})}_{\boldsymbol{\psi}}$ denote the decoder or readout head associated with conditioning signal $\mathbf{c}$. Then
\begin{equation}
    \hat{\mathbf{Y}}^{(\mathbf{c})}_{1:T}
    =
    \mathcal{D}^{(\mathbf{c})}_{\boldsymbol{\psi}}(\mathbf{F}),
\end{equation}
where $\hat{\mathbf{Y}}^{(\mathbf{c})}_{1:T}$ denotes the predicted perceptual output for task $\mathbf{c}$ over the temporal interval $1:T$. Depending on $\mathbf{c}$, this output may be a dense field, such as semantic masks, depth, normals, optical flow, or dense correspondence, or a sparse/structured output, such as pose, tracks, entities, relations, events, or uncertainty estimates.

This notation gives perception and synthesis a common conditional formulation, while preserving their different initializations,
inference regimes, readouts, and epistemic interpretations.

Perceptual inference can be probabilistic or deterministic. The probabilistic form keeps the conditional distribution $p_{\boldsymbol{\theta}}(\mathbf{F} \mid \mathbf{I}, \mathbf{c})$, which is useful when the sensory evidence is ambiguous, occluded, or underspecified. A deterministic readout instead predicts a point estimate. Let $\mathcal{G}_{\boldsymbol{\theta}}$ denote the deterministic perceptual mapping induced by the backbone and task conditioning. Then
\begin{equation}
    \hat{\mathbf{F}}
    =
    \mathcal{G}_{\boldsymbol{\theta}}(\mathbf{I}, \mathbf{c}),
    \qquad
    \hat{\mathbf{Y}}^{(\mathbf{c})}_{1:T}
    =
    \mathcal{D}^{(\mathbf{c})}_{\boldsymbol{\psi}}(\hat{\mathbf{F}}).
\end{equation}
Here $\hat{\mathbf{F}}$ is a point estimate of the output token field. This distinction is practically important. Diffusion-based synthesis may require $R \approx 40$--$50$ iterative denoising steps in contemporary video models, with each step processing a large spatio-temporal token field. Dense perception, by contrast, can often be exposed as a single forward readout from the temporally conditioned latent state. 

Equations (\ref{eq:synthesis_token_distribution}) and (\ref{eq:perception_token_distribution})  share a conditional token-field form but differ in initialization, inference, and readout. Synthesis typically uses $\mathbf{I}^{(0)}$ as noisy initialization and iterative sampling; perception encodes observed evidence as  $\mathbf{I}=\mathcal{E}_{\boldsymbol{\phi}}(\mathbf{X}_{1:T})$ and may use either a one-step point estimate or conditional sampling for multiple hypotheses. In the latter case, $\mathbf{I}$ remains fixed as evidence while stochasticity is introduced in $\mathbf{F}$ or in additional hypothesis or query tokens; noise need not be injected into the observed video except for robustness, denoising, inpainting, or posterior inference under corrupted observations.

\paragraph{Parameter specialization.} The notation $p_{\boldsymbol{\theta}}$ above denotes the common temporal
transformer model family, rather than requiring identical final parameters in
synthesis and perception. Let $\boldsymbol{\theta}_{0}$ denote the initial, potentially random, 
transformer parameters. Large-scale self-supervised generative pretraining on image and video
corpora---at the scale of billions of images and videos for systems such as
Wan~\citep{wan2025}---produces synthesis parameters
$\boldsymbol{\theta}_{s}$. Supervised perceptual fine-tuning is initialized
from $\boldsymbol{\theta}_{s}$ and produces perceptual parameters
$\boldsymbol{\theta}_{p}$. Depending on the adaptation schedule, the pretrained encoder
$\mathcal{E}_{\boldsymbol{\phi}}$ and decoder
$\mathcal{D}_{\boldsymbol{\psi}}$ may remain frozen, be fine-tuned jointly with the diffusion transformer, or be updated in alternating phases. Task-specific perceptual decoders or lightweight
structured readouts are trained where required. The two modes share the pretrained tokenization and temporal
architecture while permitting specialization of the backbone and readouts.

The generative commitment is stronger than expressing a deterministic predictor
in conditional notation. It consists in modeling distributions over
heterogeneous sensory and structured outputs within a shared temporal token
space; supporting multiple hypotheses when evidence is incomplete; predicting
missing modalities or future observations; and allowing structured perceptual
layers to function both as inferred outputs and as controls for synthesis. A
deterministic perceptual head is a useful limiting case of this broader
conditional-generative system.

\paragraph{Location of perceptual state.} In this formulation, perceptual state is not identified with any single decoded output $\hat{\mathbf{Y}}^{(\mathbf{c})}_{1:T}$, since each output is a task-conditioned projection of the inferred scene. Nor is an arbitrary latent field $\mathbf{F}$, in isolation, sufficient to constitute world state: because $\mathbf{F}$ is conditioned on $\mathbf{c}$, different queries may expose different latent projections of the same sensory evidence. The present THFM and GenCeption instantiations
\citep{wang2026thfm,wang2026eccv} remain partial and clip-conditioned; they do
not maintain durable memory across separated encounters. Shared weights and task conditioning do not themselves ensure that successive readouts describe the same maintained hypothesis. Extending these systems requires a state-update mechanism that retains entity identity and links predictions across queries and observation windows.

A future instantiation could augment these clip-conditioned representations
with recurrent state, persistent query tokens, or external episodic memory,
but these are architectural possibilities rather than capabilities of the
current motivating systems.

\paragraph{Perception-guided fine-grained control of synthesis.}
The structured variables exposed by perception can also provide fine-grained
controls for generation. Dense correspondence, geometry, pose, articulation,
segmentation, or object identity, offer substantially
more precise interfaces than language prompts alone. They could condition the
synthesis of temporally consistent digital actors \citep{lu2025gas,corona2025vlogger} controlled human--object
interaction, viewpoint and camera trajectories, scene relighting, object or
material replacement, and targeted editing that preserves unaffected scene
structure. This creates a reciprocal relationship: generative pretraining
provides priors for perception, while perceptual readouts provide spatial,
temporal, and semantic controls for fine-grained synthesis.

\paragraph{Empirical instantiation without additional self-supervised pretraining.}
In the systems motivating this paper, we do not retrain the backbone self-supervised. 
The pretrained video generator already provides strong priors for motion, temporal coherence, and cross-modal correspondence, and supervised fine-tuning is used to expose structured perceptual readouts. 
Other instantiations of the framework may combine this strategy with additional self-supervised or multimodal pretraining.

\paragraph{Dense and sparse structured prediction.}
Structured perceptual outputs differ in how they are aligned with the sensory stream. Some outputs are naturally \emph{dense}: they are defined over the same spatial or spatio-temporal support as the input video. Semantic masks, depth, surface normals, optical flow, dense correspondences, dense pose, contact maps, and per-pixel uncertainty can be represented as image- or video-like fields. These outputs can often be generated by reusing the latent video grid and decoding it through the existing RGB decoder, an adapted structured decoder, or a lightweight projection into the corresponding target space. Dense prediction preserves the spatial layout of the original observation and is therefore well suited to latent-token or ambient pixel-space supervision.

Other outputs are \emph{sparse}: they are not defined at every pixel, and their cardinality may vary across scenes and time. Examples include people, animals, objects, body parts, skeletons, object poses, tracks, relations, kinematic parameters, and interaction graphs. These variables require the model not only to decode a value, but also to decide which entities or structures should exist. For this reason, sparse prediction benefits from learned query tokens, persistent state tokens, or dynamically allocated entity tokens. Such tokens attend to the dense sensory representation, aggregate evidence over space and time, and instantiate hypotheses about entities, parts, tracks, or events when the accumulated evidence supports their existence.

This distinction matters because dense and sparse outputs expose complementary aspects of world state. 
Dense fields provide local evidence; sparse tokens can support persistence, identity, and relational structure. 
A general temporal foundation model should support both, with a shared latent representation allowing dense evidence to update sparse state and sparse state to regularize dense prediction.

\paragraph{Training regimes and loss domains.}
We consider three complementary supervision regimes that differ in where losses are applied and which decoders are active:

\begin{enumerate}[label=\roman*.]
    \item \textbf{Latent-token supervision with encoders/decoders frozen (diffusion-native). }
    The video VAE encoder/decoder are frozen; only the diffusion transformer is fine-tuned.  
This regime is most natural when structured targets can be encoded in a form compatible with the pretrained tokenizer, or when a lightweight target projection aligns them to the latent space. 
For categorical maps or metric fields, care is needed because an RGB-trained VAE may distort discrete boundaries or metric values. 
    The diffusion transformer is trained with the standard denoising loss on these latents, treating semantic/geometry video exactly as photorealistic video.  
    This preserves the pretrained latent manifold, avoids catastrophic drift, and yields strong results even with limited supervised data.

\item\textbf{Ambient pixel-space supervision via adapted decoders.}
Predictions are decoded into image-aligned target spaces, using either the existing decoder when the target can be represented as an image-like field, or lightweight adapted readout heads for categorical or metric outputs. Pixel-level losses are applied on \emph{semantic or geometric maps} (cross-entropy for segmentation, $\ell_1$ or angular losses for depth/normals).  
    This directly supervises decoded semantic or geometric predictions in the image domain.

    \item \textbf{Auxiliary ambient decoders for non--pixel-aligned outputs.}  
    For structured quantities not naturally defined per pixel (2D/3D pose, object pose, joint angles, kinematic parameters), small auxiliary decoders map diffusion latents to appropriate ambient outputs (keypoints, skeletons, parameter vectors).  
    Supervision occurs directly in ambient space (heatmaps, regression, or reprojection consistency).  
    This complements (i)--(ii) and extends supervision to abstract, spatially aggregated variables.
\end{enumerate}

\paragraph{Fine-tuning schedules: transformer-only, joint, or alternating.}
The three regimes above can be realized under different optimization schedules:

\begin{itemize}
    \item \textbf{Transformer-only fine-tuning.}  
    Keep both encoder and decoder frozen and update only the diffusion transformer.  
    This preserves the pretrained latent manifold, avoids catastrophic drift, and yields strong results even with limited supervised data.

    \item \textbf{Joint fine-tuning.}  
    Update encoder, diffusion transformer, and decoder together under mixed semantic and geometric supervision.  
    This allows the latent tokenizer and decoder to adapt to structured modalities, potentially improving spatial detail but requiring smaller learning rates and careful regularization.

    \item \textbf{Alternating or staged fine-tuning.}  
    First fine-tune the diffusion transformer with the encoder/decoder frozen (stabilizing semantics and temporal coherence), then unfreeze and lightly fine-tune the encoder or decoder in short alternating phases.  
    This gradually extends adaptation to the encoder and decoder; its benefit over joint fine-tuning requires empirical comparison.
\end{itemize}

All three schedules are compatible with the loss regimes above.  
Transformer-only fine-tuning corresponds naturally to latent-space losses (i); joint fine-tuning benefits ambient-space objectives (ii); and alternating strategies may be useful when combining ambient supervision with non--pixel-aligned decoders (iii).

\paragraph{Tokenization and multimodal extension.}
Existing diffusion generators tokenize each modality separately (video via a VAE, text via a transformer, audio as separate tokens) and align them through cross-attention.  
Our framework is directly extensible to joint audio–visual modeling through temporally aligned, modality-specific tokens, allowing acoustic events to constrain motion, geometry, and event structure without treating the modalities as interchangeable.  
Such coordinated tokenization could reduce cross-modal lag and strengthen physical grounding without modifying the diffusion loop.

\paragraph{Fusion mechanisms and structured latent state.}
The reference architecture does not require a single fusion mechanism. Different forms of fusion are appropriate for different levels of perceptual structure. A minimal implementation follows current diffusion video models and injects non-visual modalities through cross-attention: video tokens provide the dense spatio-temporal substrate, while text, audio, or task tokens condition the denoising process. This is attractive because it preserves the pretrained video manifold and allows structured perceptual outputs to be learned with limited adaptation.

A alternative implementation uses early or mid-level token fusion. Here, audio, video, language, and structured target tokens are projected into a common temporal coordinate system before several layers of joint attention. Audio tokens can be aligned to frames or sub-frame intervals; language tokens can specify tasks, entities, attributes, or relations; geometric and semantic tokens can act as target queries. Such fusion is useful when cross-modal evidence should modify the internal representation itself, not merely condition the final decoding. For example, synchronized sound may help localize an event source, language may disambiguate an entity or relation, and geometry may constrain segmentation boundaries.

A third implementation introduces persistent state or query tokens. Dense video tokens represent local evidence, while a smaller set of learned or dynamically allocated tokens maintains hypotheses about entities, parts, tracks, bodies, objects, or events. These tokens can attend to the dense sensory stream and are updated over time, functioning as a structured working state for the current world hypothesis. This is particularly useful for long-range identity, occlusion, articulated motion, interaction, and open-world entity discovery. It also creates a natural interface to structured decoders: the same persistent tokens can be decoded into masks, correspondences, pose, shape, event relations, uncertainty, or language descriptions.

These mechanisms are not mutually exclusive. A practical temporal foundation model may combine cross-attention for lightweight conditioning, joint token fusion for multimodal grounding, and persistent query/state tokens for temporally stable world structure. The architectural commitment is to a shared latent state in which heterogeneous evidence can constrain common world variables; the fusion operator may vary. This distinguishes conditional generative perception from simple multimodal prompting: the goal is to build and maintain a structured, queryable, temporally coherent representation of the perceived world.

\paragraph{Memory, retrieval, and continual update.}
A practical system may combine several forms of memory. Recurrent
working state can carry the current world hypothesis across adjacent observation
windows, using state tokens or segment-level recurrence
\citep{dai2019transformerxl}. An external episodic store can retain selected
observations and structured hypotheses for later content-based retrieval;
differentiable read--write memory and retrieval-augmented architectures provide
general precedents for such mechanisms
\citep{graves2014neural,graves2016hybrid,lewis2020retrieval}.
Recurring and independently supported structure may then be consolidated into
semantic memory, such as reusable prototypes, ontology entries, or model
parameters, while replay or parameter-stability mechanisms limit catastrophic
forgetting
\citep{kirkpatrick2017overcoming,rolnick2019experience}.

In an open-world perceptual system, an uncertain observation should therefore
not immediately rewrite the ontology. It may first revise or branch the current
working hypothesis, be retained episodically together with its provenance and
uncertainty, and become durable semantic knowledge only when recurrence,
cross-modal support, or predictive usefulness justifies consolidation.

\paragraph{Physical constraints, perceptual structure, and discovery.}
Classical vision has often used physics to make perception tractable \citep{banerjee2024physics,gartner2022differentiable, andriluka2024learned}.
Rigidity, inertia, contact, or articulation limits can regularize otherwise ill-posed inverse problems: known physical constraints can stabilize fine-tuning, improve geometric and dynamic plausibility, and reduce inconsistent perception, particularly under high uncertainty.

Physics can both constrain perception and emerge from it as stable regularities are discovered.
A system that can discover persistent entities, trajectories, and recurring interactions is already organizing observation into variables from which physical regularities can be inferred. 
Historically, scientific inquiry often proceeded by first stabilizing perceptual descriptions of objects and transformations through rigorous sampling and precise instrumental measurement—for example, using telescopes or microscopes—before expressing their regularities as quantitative laws. Ultimately, the goal is to build models that expose perceptual structure well enough for physical hypotheses, predictive laws, and counterfactual models to be learned, tested, and refined from observation. 
Known physics can regularize perception; sufficiently robust perception can also become a route to physics.

\paragraph{Relation to causal representation learning.}
Causal representation learning asks whether high-level variables and mechanisms
can be recovered from low-level observations in forms that support
generalization across environments, interventions, and distribution shifts
\citep{pearl2009causality,scholkopf2019causality}. This is closely related to
the longer-term goal of exposing entities, states, interactions, and dynamics
from temporal sensory data. The connection should not, however, be
over-interpreted: temporal prediction, structured correspondence, or
multi-output consistency do not by themselves identify causal variables or
causal mechanisms. Interventional evidence, environmental changes, action, or
additional structural assumptions may be required. The present framework
therefore treats causal structure as a possible extension of temporally
grounded perceptual state, not as an established property of THFM, GenCeption \citep{wang2026thfm,wang2026eccv}
or current video-generative backbones.

\paragraph{Relation to Video-JEPA and diffusion generators.} Video-JEPA and its recent successors~\citep{bardes2024vjepa,assran2025vjepa2,murlabadia2026vjepa21} learn predictive representations by forecasting masked or future latent embeddings from context embeddings, rather than reconstructing pixels directly. 
This reflects an important principle: for inference and action, a model should not spend all of its capacity predicting high-frequency sensory detail that is irrelevant to the task or intrinsically uncertain given the available evidence. 
At the same time, moving prediction into latent space does not by itself remove uncertainty: hidden causes, occluded objects, unobserved viewpoints, and action-dependent futures remain uncertain whether represented in pixels or in learned embeddings. 
Diffusion video generators such as Wan and Veo~\citep{wan2025,google2025veo3} also operate primarily in compressed latent spaces, although their training, decoding, post-training, and evaluation remain tied in various ways to ambient video quality.

Our framework supports latent-space and ambient/structured objectives for both perception and generation; our motivating work uses both forms of supervision.
Latent objectives can encourage compact, temporally predictive representations; ambient and structured objectives can ground those representations in observable variables and make them evaluable. 
Crucially, avoiding unnecessary pixel prediction should not be confused with avoiding spatially detailed perception. 
Many perceptual variables needed for inference and action---segmentation, depth, surface normals, or optical flow---are naturally dense and often pixel-aligned. 
Their spatial detail may be essential for manipulation, navigation, reconstruction, embodiment, human--object interaction, or scientific measurement, even if not every downstream task requires the same granularity.

Moreover, the computations required for high-quality video generation and fine-grained dense perception may be more closely related than a simple pixel-versus-latent distinction suggests. 
Our experiments indicate that a generatively pretrained temporal backbone can expose fine spatial and temporal perceptual detail, whereas the same architecture trained from scratch for structured perception does not acquire the same accuracy or transfer under limited, imperfect synthetic supervision. 
This suggests that some of the detail learned for video synthesis---appearance boundaries, motion continuity, occlusion, surface persistence, and fine geometric structure---transfers toward detailed spatial perceptual quantities, even when such detail is absent or weakly specified in the supervised perception labels.

\subsection{Empirical Signals Supporting the Hypotheses}
\label{sec:empirical_signals}

THFM ~\citep{wang2026thfm} and GenCeption ~\citep{wang2026eccv} provide empirical support for several components of the framework. THFM establishes the human-centered case through extensive evaluation against specialist systems; GenCeption provides the broader generalist instantiation through within-model ablations and comparisons with specialist, generalist, and JEPA-based systems. The following observations map these results to \emph{H1–H4}, while ontology growth, durable memory, calibrated uncertainty, full multimodal binding, and causal intervention remain open requirements.  The empirical signals can be summarized as follows:

\begin{itemize}

\item \textbf{Synthesis-oriented temporal pretraining appears central} \emph{(H1, H2).}
When the model is initialized randomly and trained directly for structured perception, the resulting system provides little evidence of comparable emergence or transfer.
By contrast, a pretrained temporal generative backbone already contains priors over appearance, motion, persistence, occlusion, articulation, and scene continuity that can be exposed through supervised perceptual readouts.
This suggests that video synthesis pretraining can internalize structure that later becomes available for perception. The comparison with training from scratch supports the value of pretraining; isolating the contribution of the generative objective requires comparisons with alternative pretraining objectives under comparable data and compute.

\item \textbf{Limited supervised synthetic data can induce broad cross-category generalization} \emph{(H2, H3).}
In the relevant experiments, supervision is limited, synthetic, and concentrated primarily on humans and human-centric structured outputs.
Nevertheless, the resulting temporal model generalizes to a broader range of categories, including animals, biological forms, articulated objects, and man-made structures.
This suggests that the transferred object is not a fixed human skeleton or a closed semantic ontology.
Instead, the model appears to reuse more general temporal and geometric priors: surface persistence, local rigidity, part-relative motion, limb-like articulation, deformation, contact, occlusion, and reappearance.

\item \textbf{Structured outputs can exceed the detail present in the supervised training signal} \emph{(H1, H3, H4).}
Even within human-centric examples, the level of fine detail recovered by the model can go beyond the detail explicitly provided by synthetic supervision.
For example, fine structures such as hair and clothing detail may be recovered despite not being explicitly represented at the same level of detail in the synthetic labels. 
This suggests that supervised perceptual fine-tuning exposes and aligns structure already active in the pretrained temporal foundation backbone.

\item \textbf{Temporal modeling can expose object-discovery-like structure} \emph{(H2).}
Persistent co-variation, common motion, deformation, occlusion and reappearance, and repeated interaction patterns can make entities, parts, tracks, and correspondences emerge in the model's explicit perceptual readouts, such as segmentation, dense pose, correspondence, or tracking outputs.
Often, this emergence is stronger than vague attentional saliency: multiple persistently moving objects appear as stable structures, either as standalone entities or as parts of an interacting configuration.
An object, part, or correspondence can therefore become visible (“pops up”) as a temporally stable perceptual output rather than as a pre-initialized structure or a directly supervised label.
This supports the idea that temporal prediction can act as a mechanism for discovering object structure, even when the underlying latent organization is only indirectly accessible through the decoded readouts.

\item \textbf{Non-trivial completion occurs through occlusion and event continuity} \emph{(H1, H2).}
The model can maintain hypotheses about identity, pose, correspondence, and likely continuation when entities disappear, reappear, or are only partially visible.
This behavior is central to perception: the relevant world state must persist beyond the currently visible pixels.
It also distinguishes temporal perceptual models from frame-level predictors, which may produce accurate local estimates while failing to maintain long-range identity or event structure.

\item \textbf{A shared temporal backbone can support many structured outputs, with both positive transfer and benchmark-dependent interference} \emph{(H1, H4).}
The GenCeption results show that a single video-generative backbone can match or surpass specialist models across several dense and sparse perceptual tasks, supporting the idea that computation can be carried by a shared substrate. 
At the same time, the ablations show that joint training is not (yet) uniformly beneficial under current metrics: some tasks improve, some remain stable, and sparse coordinate-style outputs can interfere with dense prediction.  These results provide qualified support for the representation-sharing component of \emph{H4}; whether cross-output agreement yields calibrated reliability signals remains open. Distinguishing representational interference from limitations
of task format, token design, loss domain, data mixture, or evaluation fidelity
requires controlled ablations. Correlated outputs may encourage a shared
representation, but positive transfer remains an empirical outcome of joint prediction.

\end{itemize}

\subsection{Capabilities, gaps, and integration pathways}
\label{subsec:capabilities}

The gaps identified in \S\ref{sec:why_perception_unsolved} can be organized as integration targets rather
than as an independent taxonomy. Some concern the perceptual substrate itself:
multimodal binding, persistent world state, geometric grounding, temporal
prediction, and uncertainty. Others concern extensions beyond current passive
perception: open-world adaptation, semantic and social interpretation,
counterfactual reasoning, and eventually action-oriented sensing. Table~\ref{tab:capabilities} summarizes these capabilities, their present degree of support, and plausible integration pathways, many of which remain prospective.

\begin{table}[!t]
\centering
\footnotesize
\setlength{\tabcolsep}{3.5pt}
\renewcommand{\arraystretch}{1.08}
\begin{tabularx}{\linewidth}{@{}p{2.9cm} p{2.5cm} p{1.35cm} X@{}}
\toprule
\textbf{Gap addressed} &
\textbf{Capability} &
\textbf{Status} &
\textbf{Integration pathway within conditional generation}
\\
\midrule

Fragmented sensory streams and output spaces &
Multimodal integration &
\statusPartial &
Joint audio--video tokenization, shared encoders, bidirectional prediction,
and reconstruction of missing modalities.
\\

Lack of persistent explicit state within an observation episode &
Structured working state &
\statusPartial &
Recurrent state or query tokens updated across processing windows, with
identity and evidence links maintained as context evolves.
\\

No durable memory across separated encounters &
Episodic and semantic memory &
\statusNone &
Selective storage and retrieval, provenance-aware episodic memory, and gradual
consolidation with mechanisms for controlling interference and forgetting.
\\
Weak long-horizon temporal persistence&
Long-horizon temporal prediction&
\statusPartial &
Long-horizon prediction, counterfactual rollouts, temporal contrast, and
intervention-style conditioning.
\\

Insufficient metric and physical grounding &
Geometric and physical grounding &
\statusPartial &
Depth, normals, correspondence, pose, contact, deformation, and physically
informed consistency losses.
\\

Closed-world categories and brittle transfer &
Generalization and adaptivity &
\statusPartial &
Parameter-efficient adaptation, uncertainty-driven novelty clustering, and
open-world ontology growth.
\\

Limited transition from reconstruction to interpretation &
Semantic and social understanding &
\statusNone &
Language-grounded behavioral models and narrated interaction data for intent,
communication, agency, and social cues.
\\

Prediction and synthesis not tied to explicit state &
Predictive and generative competence &
\statusPartial &
Goal-conditioned, counterfactual, and missing-modality generation for
prediction, imagination, completion, and planning.
\\

Uncalibrated or inconsistent outputs &
Self-consistency and uncertainty &
\statusPartial &
Sampling variance, ensemble heads, cross-modal disagreement, consistency
checks, and confidence-aware abstention.
\\

Dependence on exhaustive dense annotation &
Data-efficient structured adaptation &
\statusFull &
Synthetic supervision, temporal pretraining, shared structured outputs, and
transfer from limited task-specific labels.
\\

Extension beyond passive perception &
Action-oriented perception &
\statusNone &
Task-directed and information-seeking action through policy or attention
mechanisms that select viewpoints, sensors, queries, or interventions using
the current perceptual state, uncertainty, and predicted outcomes.
\\

\bottomrule
\end{tabularx}
\caption{
Capabilities, unresolved gaps, current support, and integration pathways for
open-world temporal perception. The table connects the earlier diagnosis of
why perception remains unsolved to concrete mechanisms within a
conditional--generative framework.
Legend: \statusFull~demonstrated in the motivating systems or closely related
work; \statusPartial~partially demonstrated or plausible but incomplete;
\statusNone~future extension.
}
\label{tab:capabilities}
\end{table}

\subsection{Evaluation Beyond Benchmarks}
\label{subsec:evaluation}

Evaluating open-world perception requires both task-specific accuracy and
structural diagnostics. Standard measures---including segmentation accuracy,
depth and normal error, pose accuracy, and tracking precision---remain
essential for grounding comparison with specialist systems. However, they
capture only fragments of the broader objective: whether a model maintains
coherent, temporally stable, and physically meaningful structure.

Exhaustive dense annotation is unavailable for many organisms, artifacts,
mechanisms, materials, and long-tailed interactions. Synthetic data provide
controlled variation and known geometry or dynamics, but can reproduce
simulator assumptions and omit precisely the structures that are poorly
understood. Evaluation should therefore combine reliable real benchmarks,
controlled synthetic tests, temporal and multi-view consistency,
reconstruction fidelity, uncertainty calibration, independent-model
agreement, and targeted expert review. These diagnostics complement rather
than replace ground truth and must themselves be tested for shared bias or
coherent hallucination.

\vspace{2mm}
\noindent
When exhaustive annotation is unavailable, evaluation must additionally probe coherence, uncertainty, emergence, cross-modal agreement, and, where appropriate, causal structure. We consider the following quantitative and diagnostic tests:

\begin{itemize}

\item \textbf{Temporal persistence, memory, and evidence-driven revision.}
Test whether perceptual states persist beyond the immediate context and remain
recoverable and revisable as evidence accumulates. Controlled delayed-recall
sequences can introduce an entity or event, remove it for progressively longer
periods with intervening distractors, and reintroduce it under changed
appearance, viewpoint, or sensory evidence. Measure identity and state recovery
as a function of delay, interference, and model updates. Confirming or
contradictory evidence should reveal whether the system preserves, revises,
branches, or invalidates a stored hypothesis rather than silently replacing it.
Such protocols can draw on precedents from long-term tracking, segment-level recurrence, external memory, and experience replay 
\citep{fan2019lasot,dai2019transformerxl,graves2016hybrid,
rolnick2019experience}.

\item \textbf{Evaluation without exhaustive human annotation.}
When explicit ground truth is unavailable, evaluate agreement among
independently trained models, consistency across augmentations and viewpoints,
temporal stability of discovered entities, cycle consistency of dense
correspondences, compatibility between depth and normals, and reconstruction or
re-synthesis fidelity. Human assessment can then provide targeted validation of
representative successes and failures rather than serving as the sole scalable
source of ground truth.

\item \textbf{Cross-modal binding and synchrony.}
Test whether visual, auditory, textual, haptic, and proprioceptive observations
are bound to the same entities and events at the appropriate times. Cross-modal
retrieval, temporal alignment error, event correspondence, and sensitivity to
controlled temporal or semantic mismatches can reveal incorrect binding,
desynchronization, and dominance or neglect of individual modalities.

\item \textbf{Geometric and physical stability.}
Evaluate whether predicted depth, normals, shape, contact, and motion remain
consistent across viewpoint changes, occlusion, and temporal continuation.
Multi-view reprojection error, photometric cycles, normal--depth compatibility,
contact consistency, and physical motion constraints provide quantitative
probes of internal 3D and physical coherence. Controlled changes in viewpoint,
lighting, and occlusion can further test whether the model preserves underlying
structure rather than confusing appearance variation with physical change.

\item \textbf{Open-world novelty, abstention, and ontology growth.}
Test whether the model recognizes novel phenomena and abstains selectively when its current concepts are inadequate. Risk–coverage curves \citep{geifman2019selectivenet} measure selective prediction and abstention, while detection performance on controlled novelty sets \citep{bendale2016towards,hendrycks2021natural} measures novelty recognition. Candidate concepts should be tested on subsequent observations: whether they support recognition or prediction beyond the examples that formed the cluster, and whether they do remain distinct under changes in viewpoint, appearance, and context.

\item \textbf{Uncertainty as a predictor of error.}
Determine whether internal uncertainty predicts actual perceptual or
reconstruction error wherever ground truth or self-consistency tests are
available. Sample dispersion, Monte Carlo dropout, diffusion-sample variance, and
disagreement among independent decoders provide intrinsic uncertainty
estimates. Correlation with observed error, reliability diagrams, and
calibration error then measure whether these signals predict model reliability
rather than merely reflecting internal variability.

\item \textbf{Interpretable emergence and representational structure.}
Analyze whether functional structure emerges within the model: attention heads
specializing in parts, slots representing entities, persistent tokens tracking
objects, or decoder branches aligning with articulated structure.
Representational similarity analysis, latent-space clustering, feature
attribution, and interventions on entity or state tokens can quantify these
patterns and expose the organization underlying perceptual inference.

\end{itemize}

\noindent
Together, these axes test whether task accuracy is supported by a coherent and revisable account of world structure.

\subsection{Resources and Scaling}
\label{subsec:resources}

Scaling remains central, but the relevant question is whether explicit
perceptual structure becomes more coherent, transferable, and calibrated as a
function of model capacity, compute, data diversity, temporal context, modality
breadth, and structured supervision. Suitable studies should measure object
persistence, cross-category correspondence, geometric fidelity, multi-output
consistency, uncertainty calibration, data efficiency, and open-world
discovery. The GenCeption results provide an early instance of this analysis,
but systematic perceptual scaling laws remain to be established. A complementary systems-level test is whether a shared temporal substrate
reduces adaptation data, duplicated computation, and integration cost relative
to collections of specialist models at matched accuracy, latency, and
reliability.

\subsection{Scope and Caveats}

The framework distinguishes computational, algorithmic, and implementation levels \citep{marr1982vision}, together with four cross-cutting hypotheses. Computationally, generalist open-world perception must exhibit the properties identified in \S\ref{sec:why_perception_unsolved}, maintaining perceptual information in a persistent, queryable state that binds heterogeneous evidence across time and exposes compatible structured readouts. At the algorithmic level, conditional generative perception specifies one route toward this goal: temporal evidence conditions a shared model from which structured perceptual readouts are inferred, deterministically or probabilistically. Within this route, generative pretraining for photorealistic synthesis provides the temporal substrate subsequently adapted for perception. At the implementation level (\S\ref{subsec:architecture}), synthesis-pretrained temporal backbones, with particular tokenizers, transformer architectures, decoders, readout heads, and adaptation schedules, instantiate this route. The four hypotheses (\S\ref{subsec:hypotheses}) are cross-cutting claims about model properties that can be empirically tested and potentially falsified. This separation also yields several caveats.

First, generative priors are useful only insofar as they remain constrained by evidence. 
The same prior that supports completion through occlusion, missing modalities, or sparse supervision can also hallucinate structure, over-regularize rare cases, suppress genuinely novel phenomena, or impose familiar dynamics where the observations are ambiguous. 
Photorealistic or temporally smooth generation is therefore not sufficient evidence of correct perceptual structure. 
A generative perceptual model must be judged by whether its inferred geometry, entities, correspondences, events, affordances, and uncertainties remain stable under independent evidence, perturbations, and downstream use.

Second, observational temporal data can reveal substantial world structure even without active control by the model. 
Video may contain moving viewpoints, moving agents, occlusion and reappearance, contact, manipulation, deformation, object use, social interaction, and many other consequences of action. 
A temporal generative model trained on such data can learn rich priors about persistence, geometry, interaction, and change from observation alone. The limitation is evidence selection: the model cannot choose the viewpoint, intervention, contact event, or counterfactual variation needed to resolve a particular ambiguity. 
Some ambiguities may therefore remain unresolved when the observations lack the discriminating evidence. 
In such cases, active sensing and exploration, as discussed in \S\ref{sec:levels-of-observation}, are mechanisms for acquiring additional evidence targeted to unresolved hypotheses, not replacements for temporal perception.

Temporal prediction, interventional prediction, and counterfactual inference
should consequently be distinguished. Observed temporal regularities support
predictions under familiar trajectories; intervention data reveal how the
system responds when actions or conditions are changed; counterfactual
inference additionally requires reasoning about alternatives relative to a
particular observed history. A temporal generative backbone may support all
three, but observational coherence alone establishes only the first.

Third, tightly specified applications may use implicit sensory-to-action mappings effectively. Explicit state becomes most valuable when the system must generalize, communicate, diagnose failures, combine tasks, support reasoning, remain auditable, or operate under uncertainty. The relevant design choice is application-dependent: generality, interpretability, data efficiency, compute cost, and accuracy must be traded against the simplicity and reliability of specialized systems.

Fourth, open-world ontology growth requires more than uncertainty. 
A high-uncertainty prediction may reflect sensor noise, occlusion, rare pose, domain shift, annotation ambiguity, or a genuinely new category, part, state, or event. 
A new perceptual concept should be stabilized through recurrence, persistence across observations, predictive usefulness, cross-modal support, and, where appropriate, language, expert validation, or interaction. 
Ontology growth should be treated as the formation of reusable perceptual structure, not merely as clustering residual errors.

Fifth, open-world perception requires memory at multiple time scales: short-term state for tracking, occlusion, and event continuity; episodic memory for places, individuals, situations, and repeated encounters; and semantic memory for categories, mechanisms, affordances, and failure modes accumulated across experience. 
The temporal token fields and state/query tokens discussed above provide a possible substrate for such memory, but the mechanisms for durable storage, retrieval, consolidation, forgetting, and revision remain open architectural questions.

Finally, internal consistency diagnostics may preserve shared biases or support coherent hallucinations. Open-world evaluation must combine them with curated benchmarks, synthetic stress tests, controlled perturbations, independent measurements, expert review, and explicit failure-seeking protocols.

\section{Applications and Societal Impact}
\label{sec:impact}

A generalist open-world perceptual model is most valuable in settings where
isolated predictions are insufficient and an intelligent system must maintain,
communicate, revise, or act upon an explicit account of the world. Persistent
entities, geometry, events, relations, and uncertainty provide interfaces for
reasoning, diagnosis, human oversight, and safe interaction that cannot be
replaced by a collection of unrelated task outputs.

In robotics and autonomous sensing, such state could support manipulation,
navigation, interaction, and decisions about when additional evidence is
required. In graphics, virtual production, and augmented or virtual reality, structured perceptual state could support reconstruction and motion capture
while providing geometry-, correspondence-, pose-, contact-, and
camera-conditioned controls for digital actors, AI-assisted cinematography,
telepresence, and immersive interaction beyond what language prompts alone can
specify. In medicine and science, structured temporal inference
could help interpret surgical video, physiological signals, cell dynamics,
ecological behavior, material processes, or other observations for which
task-specific annotation is limited but persistent structure is scientifically
meaningful. The same substrate could support accessibility applications such as
sign-language understanding, physical gesture interpretation, and multimodal
interfaces adapted to different bodies and sensory capabilities.

Because perception and synthesis share model components and representational
structure, these applications also create substantial risks of fabricated or
misrepresented evidence. Generated or reconstructed content should therefore be
accompanied by provenance, uncertainty, and a distinction between observed,
inferred, and synthesized structure. The societal value of these systems will
depend not only on what they can reconstruct or generate, but on whether their
claims about the world remain inspectable, calibrated, and accountable.

\section{Perception as Foundation and Frontier of Superintelligence}
\label{sec:perceptual-superintelligence}

\paragraph{AGI, ASI, and the missing perceptual dimension.} Recent work on artificial general intelligence and artificial superintelligence (AGI and ASI,
respectively) emphasizes that intelligence should not be treated as a single
threshold. The ``Levels of AGI'' framework distinguishes performance,
generality, and autonomy: systems may be narrow or general, may range from
emerging to competent, expert, virtuoso, or superhuman performance, and may be
deployed as tools, consultants, collaborators, experts, or autonomous agents
\citep{morris2024levels}. Recent work on the transition from AGI to ASI extends
this framing beyond individual human performance: ASI is discussed relative to
the capabilities of expert groups, institutions, large human organizations, and
eventually systems that may accelerate scientific and technological progress
through scaling, paradigm shifts, recursive improvement, or multi-agent
collectives \citep{genewein2026agi2asi}.

These frameworks are useful, but they largely leave aside the developmental
and world-modeling role of perception in general intelligence. As discussed in \S\ref{sec:miss-gpt}, frontier systems are already multimodal: they analyze images, diagrams, charts, and video together with language-based reasoning, and some AGI taxonomies include visual or embodied tasks. Operational discussions of AGI and ASI nevertheless foreground language, mathematics, problem solving, games, and scientific reasoning, usually treating perception as another capability domain or benchmark. The missing dimension is a developmental and world-modeling account of how an intelligent system acquires, maintains, and revises structured knowledge of a world \citep{morris2024levels,genewein2026agi2asi}. A fuller account of general intelligence concerns both what a system can do with available representations and how it forms the representational resources through which it reasons, constructs analogies and hypotheses, imagines alternatives, and relates them back to the world.

\paragraph{Developmental grounding.} For organisms, intelligence is closely tied to adaptive regulation under uncertainty:
maintaining viability, acquiring resources, avoiding threats, coordinating
with others, and adjusting behavior as conditions change. Perception supplies
much of the evidence on which such regulation depends by identifying what is
present, what has changed, what is relevant, and which actions remain
possible. Developmentally, inherited neural organization provides substantial
priors, but does not by itself constitute knowledge of an external world. Patterned sensory experience calibrates perceptual systems \citep{blakemore1970development}. More broadly, sensorimotor experience and interaction with the environment contribute to the development of grounded conceptual representations \citep{smith2005development,
barsalou2008grounded}. Perceptual development is therefore part of the process
through which inherited structure informs understanding,
imagining, and acting in a world. Language,
culture, instruments, and institutions subsequently allow cognition to extend
far beyond immediately available sensory evidence and beyond the experience of
a single individual.

Higher cognition is not tied to any single sensory modality. Sensory loss can be accompanied by substantial behavioral compensation and cross-modal neural reorganization, although the effects depend on the affected modality and the timing and extent of deprivation \citep{frasnelli2011crossmodal}. The relevant prerequisite is sufficiently rich and structured coupling through which the world can constrain learning, irrespective of the particular sensory channels available.

Developmental grounding should not be understood as mechanically determining the content or limits of thought. General intelligence is not exhausted by maintaining an accurate sensory-based model of the actual world; it also constructs possible worlds, formal systems, counterfactuals, analogies, and novel hypotheses that may depart radically from immediate experience. Once established, such abstract systems may be developed further and assessed primarily through internal criteria such as consistency, proof, formal consequence, explanatory coherence, or conceptual fertility, without their internal claims needing to be tested against sensory observation. In pure mathematics, for example, sensory experience ordinarily has no direct justificatory role in establishing the truth of a theorem.

Nevertheless, the primitives, transformations, and imaginative operations through which humans construct such systems arise
within cognitive lives already shaped by perception, action, language, and
social experience \citep{barsalou2008grounded}. Mathematics can describe
structures with no direct sensory counterpart, while using symbols, spatial and
quantitative intuitions, diagrams, and lower-dimensional projections to make
them cognitively accessible. These representations are not literal copies of
the abstract objects they support, and their analogies may sometimes mislead.
They are better understood as perceptual and symbolic scaffolds that enable
thought to transcend the immediately perceptible reality without becoming
developmentally independent of perception.

\paragraph{Perception in the scientific and technological loop.}
Perception is not merely an initial developmental scaffold that can be
discarded once abstract cognition has been launched. In mature intelligence,
and especially in science, medicine, and technology, it remains part of a
recurrent cognitive loop. Observation generates questions, exposes anomalies,
suggests analogies, and constrains hypotheses; abstraction reorganizes what is
subsequently sought, measured, and noticed; and further observation tests,
redirects, or extends the resulting ideas
\citep{hanson1958patterns}. Instruments do not remove perception from this
process, but expand it. Telescopes, microscopes, medical imaging, and distributed
sensors make otherwise inaccessible structure perceptually and quantitatively
available.

One useful, though necessarily schematic, vocabulary for this loop
distinguishes induction, deduction, and abduction
\citep{peirce1878deduction,simon1973discovery,zahavy2026jump}. Induction
extracts regularities from observations and includes much of statistical
learning and pattern discovery; its conclusions are supported by evidence
rather than logically guaranteed. Compression-based accounts connect this
process to curiosity and discovery: improvements in prediction or compression
can reveal previously unrecognized structure in the observations
\citep{schmidhuber2009compression}. Deduction derives consequences from
accepted assumptions and is exemplified computationally by formal theorem
proving. Abduction proposes a hypothesis under which an observation---often an
anomaly or surprise---would become expected or intelligible, a process often
characterized as inference to the best explanation
\citep{harman1965inference}. It is not arbitrary association, but neither is
it truth-preserving: competing explanations must be compared in terms of their
coverage of the evidence, coherence, simplicity, prior plausibility, causal
adequacy, and discriminating power. Manipulative abduction may additionally
involve interaction with physical, simulated, diagrammatic, or other external
representations \citep{magnani2009abductive}.

In practice, these processes are interdependent: abductive proposals are
constrained by regularities supported inductively, deduction determines what
would follow from each proposal, and new evidence discriminates among the
resulting accounts. A perceptual world model does not by itself guarantee
invention; to the extent that it exposes entities, relations, uncertainties,
anomalies, and counterfactual possibilities in a form available to reasoning,
it may nevertheless become part of the machinery through which hypotheses are
formed and evaluated. This framing connects classical accounts of inquiry to
contemporary efforts to use AI to accelerate scientific discovery
\citep{hassabis2024accelerating}.

For an abstraction to become scientifically consequential, however, it must do
more than remain internally coherent: it must specify observables, identify
measurable implications, and motivate informative interventions through which
it can be tested against the perceptual world and fitted productively to it.
Advances in physics often depend on connecting mathematical structures to
detectable phenomena; advances in medicine depend on relating hypotheses to
biological signals, anatomical structure, disease progression, and treatment
outcomes; advances in technology require abstract principles to be realized in
materials, mechanisms, environments, and human use. In each case, advanced
perception helps reveal where abstractions apply, where they fail, and how they
may be transformed into useful knowledge and physically realizable innovation.

Abstract systems may still possess coherence, beauty, depth, or intellectual
value without any known practical application, and utility is neither their
sole measure nor necessarily visible at the time of discovery. Nevertheless,
among the indefinitely many internally consistent formal systems, those that
eventually explain, predict, reveal, or positively transform aspects of the
world often acquire distinctive scientific and societal significance. The
return from abstraction to observation, intervention, and consequence, as enabled by perception, is
therefore not a universal criterion of thought, but it is central to
world-directed intelligence and to those discoveries most consequential for human progress.

\paragraph{Two questions for AGI and ASI.}
The first is foundational: what forms of perceptual development and continuing
coupling to the world are required both to furnish a general intelligence with
the resources for abstraction, analogy, imagination, and creativity and to
reconnect those capacities to evidence and consequence, when appropriate, as necessary for grounded discovery and innovation? The
second concerns perceptual competence itself: what would it mean for a system
to perceive beyond the reach of an individual human, an expert community, or
even a biological species? A system might display formidable abstract and
creative reasoning while remaining weakly grounded in sensory evidence.
Conversely, it might exceed humans in sensing, measurement, and recognition
without integrating these abilities into general cognition. Perceptual
superintelligence should therefore be understood both in relation to broader
intelligence---as a developmental and epistemic foundation and a continuing
partner---and as a capability domain in its own right.

\paragraph{Human and biological reference points.} Human perception is selective, active, embodied, and
task-dependent. Vision is foveated and attentional; memory stabilizes
perception across eye movements and occlusion; action changes the evidence
being acquired. Human perception is an adaptive process for extracting behaviorally useful structure from selective, foveated, and action-dependent evidence. Other modalities reinforce the same point. Audition is spatial, but it
organization differs from foveated vision: it is distributed, temporally
precise, and strongly source-oriented. It supports localization, event
detection, source separation, speech, rhythm, and causal inference under
ambiguity. Haptics and proprioception link perception directly to contact,
material, force, body state, and action. Superhuman perception in these
modalities would not simply mean greater sensitivity, but more reliable
decomposition of the world into sources, events, contacts, or materials.

The human reference point is pragmatic rather than absolute. Other species
exceed humans along particular sensory dimensions and inhabit different
perceptual worlds shaped by their sensors, bodies, niches, and possible
actions, as discussed in \S\ref{sec:why_perception_unsolved}. Perceptual intelligence is 
perceiver-relative even when human individuals, experts, and institutions
remain useful operational reference points for AGI and ASI.

\paragraph{Perceptual Competence Axes.} Perceptual intelligence spans several axes whose human, expert, institutional, and superhuman thresholds differ.  A system may estimate dense
geometry more precisely than a human observer, recognize more long-tailed
categories than an individual, or process vastly more video, while still
failing to bind entities through occlusion, interpret structure, or select what
matters for a task. Perception may therefore need to be evaluated along
several axes rather than by one aggregate score.

A first axis is \textbf{open-world categorical perception}. The relevant comparison includes both individual humans and expert communities or institutions that maintain taxonomies, museum collections, diagnostic criteria, field observations, databases, and scientific literature. A machine system may exceed an
individual human in recognizing long-tailed biological species, artifacts,
materials, actions, or rare events, while still falling short of the richer
structural knowledge accumulated by such institutions. Recognizing a rare
bird, insect, tool, or material is not the same as understanding its
morphology, variation, habitat, behavior, development, function, or relation
to neighboring categories. This is where perception begins to link to
cognition: sensory evidence becomes connected to concepts, taxonomies,
functions, explanations, and possible inferences.

A second axis is \textbf{dense structured perception}. Dense estimates of geometry, semantics, or correspondence may already go beyond unaided human perception. Some of these outputs are metric, others are semantic or relational,
but all are spatially precise and defined over images, videos, surfaces, or
other dense supports. Humans can reason about many of these quantities
qualitatively, and experts can measure or annotate them with tools, but no
human observer, and no human organization by manual effort alone, can produce
temporally coherent dense structured estimates over large video archives.
Along this axis, superhuman perception means spatial, temporal, and
statistical resolution beyond biological perception. This form of
superhumanity is real but narrow: it concerns the density, precision, and scale
of perceptual estimation, not necessarily the integration of those estimates
into a coherent account of the world.

A third axis is \textbf{selective integrative perception}, where biological
perception remains an important reference point. Humans are often poor dense
estimators, but strong selective integrators. They maintain object permanence,
bind parts into entities, infer causes from sparse evidence, ignore irrelevant
variation, and allocate attention to informative regions. Along this axis,
many machine systems remain less mature: they may estimate dense fields or
recognize categories, but fail to maintain a coherent account of what
persists, what changes, what is occluded, what belongs together, what is
relevant, and what remains uncertain. Human perception remains a competitive standard because it selects, binds, and interprets efficiently  with respect to most non-trivial human tasks, even without measuring everything. 

A fourth axis is \textbf{generalist open-world temporal perception}. This is the
integrative frontier targeted here. A perceptual system should combine
open-world category knowledge, dense structured evidence, temporal
persistence, multimodal binding, selective computation, and uncertainty into a
coherent model of the perceived world. It should infer persistent entities,
physical state, multimodal causes, and alternative hypotheses from incomplete
and changing evidence.

This axis also captures two forms of perceptual scale that are not available to
a single biological observer. First, artificial systems may integrate sensing
channels beyond ordinary human perception, including depth, thermal,
event-based, inertial, radar, lidar, hyperspectral, or distributed acoustic
measurements. Second, they may integrate evidence across many observers,
instruments, viewpoints, places, and times, approximating a community of
perceivers rather than one embodied viewpoint.

Classical multisensor data fusion provides a general framework for integrating evidence from multiple sensors and sources \citep{hall1997multisensor}. Modern systems instantiate this principle across configurations ranging from satellite, aerial, and ground-based sensing to autonomous vehicles combining cameras, radar, lidar, inertial measurements, maps, and observations accumulated along a trajectory. Mapping and fleet-level systems
can extend this integration across multiple vehicles, locations, and times.
These systems demonstrate that perceptual evidence need not originate from one
sensor or one embodied observer. Their sensing configurations, objectives, and
acquisition programs, however, are often largely specified in advance, even
when the resulting estimates cover substantial spatial and temporal extents.

A further step is to make evidence acquisition itself agentic. Classical active
perception already proposed that sensing should be selected according to the
task and the uncertainty of the current interpretation
\citep{bajcsy1988active,bajcsy2018revisiting}. Although this principle is not
intrinsically limited to short time scales, most classical formulations and
implementations operated through relatively local and task-bounded sensing
loops: selecting the next viewpoint or measurement, updating the current
interpretation, and repeating over a comparatively short horizon.
Contemporary agents and orchestrators, when coupled to persistent world state
and long-horizon memory, create the possibility of applying this principle
across broader spatial, temporal, and organizational scales. Extending current agentic and robotic systems in this direction could allow a system to select which sensor, observer, viewpoint, model, or tool to query; dispatch mobile platforms; change sensing position or resolution; request targeted measurements; perform interventions; assess whether an action succeeded; and replan as the shared perceptual state changes \citep{ahn2024autort,abdolmaleki2025geminirobotics15}. Multiple agents may
divide sensing tasks, compare asynchronous observations, preserve and revisit
unresolved hypotheses, resolve conflicting evidence, and allocate bandwidth,
energy, time, or risk according to expected information value. The opportunity
is to leverage classical sensor fusion or active vision principles by combining distributed passive sensing, active evidence acquisition, model and
tool selection, and physical intervention within a persistent and revisable
perceptual process extending across space and time.

This level is not simply human-like, institutional, or superhuman in one narrow
sense. It requires combining strengths that currently span 
biological perception, artificial sensing, machine-scale computation,
distributed observation, and scientific knowledge.

In this sense, progress toward perceptual superintelligence should be measured
by the ability to expose coherent world structure across category, resolution,
time, modality, action, and uncertainty. Such a substrate would link
perception to cognition: it would ground abstraction, support counterfactual
imagination, constrain planning, and connect sensory evidence to the models of
the world on which higher intelligence depends.

\section{Responsible Development and Ethical Considerations}

Responsible generative perception depends on preserving the distinction among observed, inferred, and synthesized structure. Outputs should retain provenance, uncertainty, and links to supporting evidence; underdetermined states should trigger abstention, deferral, or targeted evidence acquisition. Evaluation should report coherence, calibration, failure modes, and known limitations using auditable datasets, benchmarks, and diagnostic tools. These mechanisms can also reveal how biases in multimodal corpora propagate through generative conditioning into structured perceptual outputs. Geometry, time, and physical consistency constrain inference but cannot correct missing, selectively recorded, or socially biased evidence; dataset documentation, curation, and targeted evaluation therefore remain necessary.

\section{Concluding Perspective}

The position developed in this paper is that generalist perception constructs an explicit world state that persists through time, binds multimodal evidence into a coherent account of the world and its evolution, represents uncertainty, and remains revisable as new evidence arrives. Its perceptual and generative readouts are compatible projections of that state.

Conditional generative temporal models offer a plausible route toward this goal because synthesis-oriented temporal pretraining requires models to capture regularities in how sensory observations evolve over time. THFM and GenCeption~\citep{wang2026thfm,wang2026eccv} provide initial evidence for a significant and perhaps surprising form of perceptual emergence: structure acquired through synthesis-oriented temporal pretraining can be exposed through dense and sparse structured readouts, generalize beyond directly supervised categories, and support accurate and efficient perceptual inference. They do not determine the final implementation. Alternative predictive, recurrent, or hybrid architectures may instantiate the same broader principles, provided that they support coherent temporal state rather than only plausible sensory generation or disconnected task outputs.

The decisive questions are empirical: whether temporal pretraining
yields stable and transferable perceptual structure; whether multimodal
evidence can be genuinely bound into a coherent state; whether that state
remains calibrated and revisable under ambiguity, novelty, and distribution
shift; and whether persistence can extend beyond increasingly long but bounded
context windows into selective and durable episodic and semantic memory.

If language models made text a common interface for reasoning, temporal
perceptual foundation models may make explicit world structure a common
interface for multimodal intelligence. This interface supports connecting sensing, reasoning, simulation, and action, leading to physical AI: a shared perceptual core on which task- and embodiment-specific policies can be built. More broadly, such a substrate may help
close a missing loop in contemporary accounts of intelligence: between the
sensory evidence from which knowledge is formed, the abstractions through which
it is extended, and the observations and interventions through which it is
revised and made consequential. The goal is not only to recognize what is
present in the sensory evidence, but to infer what persists, what changes, what
remains hidden or uncertain, and what new structure the world reveals. A
perceptual foundation model should not merely label the world; it
should help discover it.

\section*{Acknowledgements}

The work underlying this paper benefited greatly from the intellectually stimulating research environment at Google and from the broader exchange of ideas across its research community. I am grateful to Misha Andriluka, Eduard Gabriel Bazavan, Andrei Zanfir, Chris Dyer, and Andrew Zisserman for sustained collaboration and contributions to the research program from which this position paper emerged. 
I thank Letian Wang, Chuhan Zhang, Rishabh Kabra, Jasper Uijlings, Steven Waslander, Joao Carreira, and Kaiming He for their collaboration and contributions to the research and companion empirical work that helped shape this position paper.
I am grateful to Richard Hartley, Bill Freeman, Vlad Olaru, Dima Damen, and David Fleet for valuable discussions, feedback, and encouragement. This manuscript also benefited from the use of generative AI tools for documentation, language editing, and iterative refinement of its presentation. The perspective, framing, scientific arguments, technical formulations, interpretations, and conclusions remain those of the author.

\FloatBarrier
\bibliographystyle{iclr2025_conference}
\bibliography{references_clean-2}

\end{document}

%% file: math_commands.tex
\usepackage{amsmath,amsfonts,bm}

\def\eqref#1{equation~\ref{#1}}

\def\1{\bm{1}}

\DeclareMathAlphabet{\mathsfit}{\encodingdefault}{\sfdefault}{m}{sl}
\SetMathAlphabet{\mathsfit}{bold}{\encodingdefault}{\sfdefault}{bx}{n}

